\documentclass[10pt,twocolumn,letterpaper]{article}

\usepackage{wacv}              

\usepackage{graphicx}
\usepackage{amsmath}
\usepackage{amssymb}
\usepackage{booktabs}
\usepackage{float}
\usepackage{multirow}

\usepackage{arydshln}
\usepackage{pifont}

\newcommand{\cmark}{\ding{51}}%
\newcommand{\xmark}{\ding{55}}%

\definecolor{wacvblue}{rgb}{0.21,0.49,0.74}
\usepackage[pagebackref,breaklinks,colorlinks,allcolors=wacvblue]{hyperref}

\def\wacvPaperID{1029} 
\def\confName{WACV}
\def\confYear{2026}

\title{VectorSynth: Fine-Grained Satellite Image Synthesis with Structured Semantics}

\author{
Daniel Cher$^{*}$ \quad Brian Wei$^{*}$ \quad Srikumar Sastry \quad Nathan Jacobs\\
Washington University in St.\ Louis\\
{\tt\small \{cher, b.j.wei, s.sastry, jacobsn\}@wustl.edu}\\
{\small $^{*}$Equal contribution}
}

\begin{document}

\twocolumn[{%
  \renewcommand\twocolumn[1][]{#1}%
  \maketitle
  \vspace{-0.6cm}
  \begin{center}
    \captionsetup{type=figure}
    \includegraphics[width=\linewidth]{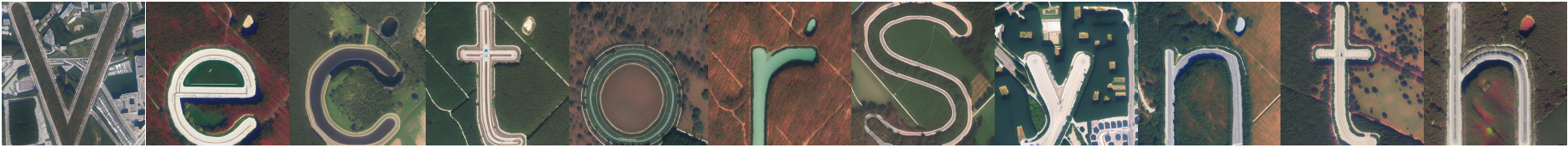}
    \captionof{figure}{VectorSynth logo synthesized using learned OSM‑based pixel
    embeddings. Each letter is generated with distinct OSM tag combinations
    (e.g.\ industrial, farmland, geological features), demonstrating VectorSynth’s
    fine‑grained semantic control over satellite‑image synthesis.}
    \label{fig:vectorsynth_logo}
  \end{center}
  \vspace{-0.4cm}
}]

\maketitle
\begin{abstract}

We introduce VectorSynth, a diffusion-based framework for pixel-accurate satellite image synthesis conditioned on polygonal geographic annotations with semantic attributes. Unlike prior text- or layout-conditioned models, VectorSynth learns dense cross-modal correspondences that align imagery and semantic vector geometry, enabling fine-grained, spatially grounded edits. A vision language alignment module produces pixel-level embeddings from polygon semantics; these embeddings guide a conditional image generation framework to respect both spatial extents and semantic cues. VectorSynth supports interactive workflows that mix language prompts with geometry-aware conditioning, allowing rapid what-if simulations, spatial edits, and map-informed content generation. For training and evaluation, we assemble a collection of satellite scenes paired with pixel-registered polygon annotations spanning diverse urban scenes with both built and natural features. We observe strong improvements over prior methods in semantic fidelity and structural realism, and show that our trained vision language model demonstrates fine-grained spatial grounding. The code and data are available at \url{https://github.com/mvrl/VectorSynth}.

\end{abstract}    
\section{Introduction}
\label{sec:intro}

\begin{figure}
  \centering
  \includegraphics[width=0.85\linewidth]{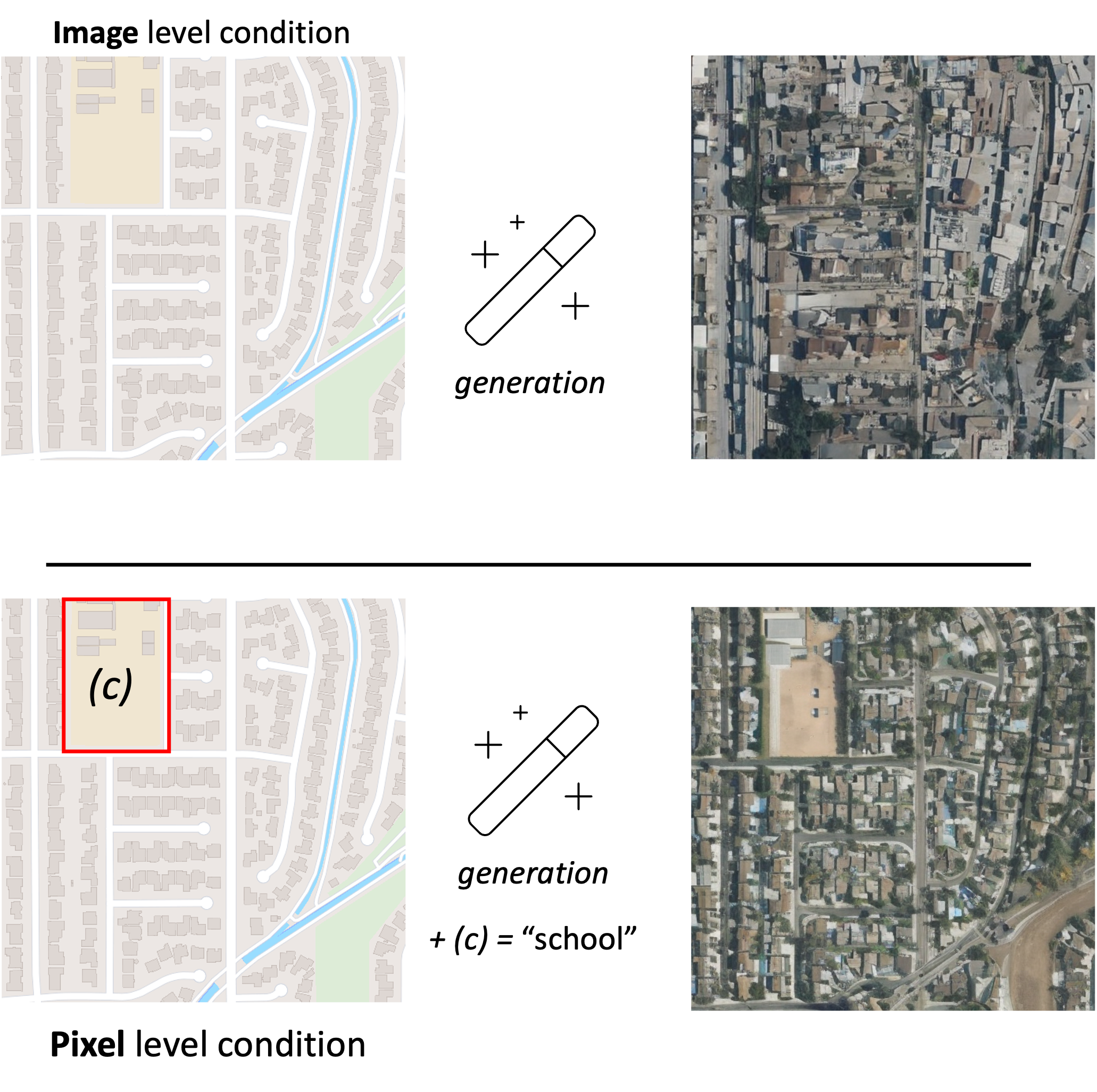}
  \caption{
  VectorSynth enables precise, fine-grained control over both spatial location and semantic content during satellite image synthesis. top: Image-level conditioning lacks the ability to target specific regions or object types. bottom: VectorSynth allows conditioning on individual polygons with semantic labels (e.g., $c$ = “school”), producing coherent imagery that respects both spatial extent and semantics.
  }
  \label{fig:task_description}
\end{figure}

Text-to-image generative models have witnessed rapid progress in recent years, driven by advances in large-scale vision-language pretraining. Models such as DALL·E~2~\cite{ramesh2022hierarchical} and Stable Diffusion~\cite{rombach2022high} demonstrate the remarkable ability to synthesize images from natural language prompts, ranging from surreal abstractions to photorealistic renderings. Beyond raw generation, these models enable downstream tasks such as inpainting~\cite{lugmayr2022repaint}, layout-to-image translation~\cite{li2023gligen}, and text-guided image editing~\cite{brooks2023instructpix2pix, kawar2023imagic}. Architectures like ControlNet~\cite{zhang2023adding} extend this further by introducing a framework to adapt these large pre-trained diffusion models for domain-specific applications. 

Recently, the remote sensing community has begun exploring generative models for various Earth observation tasks, including disaster response, environmental monitoring, poverty estimation, and urban planning, which have shown to benefit from generated satellite imagery~\cite{pang2023ssrgan, mahara2024multispectral, he2021spatial, goktepeecomapper}. Diffusion-based approaches, such as DiffUCD~\cite{zhang2023diffucd} and HySCDG~\cite{benidir2025change}, have shown that using an appropriate conditioning technique yields rich synthetic satellite data that can support downstream remote sensing tasks.

However, existing approaches for satellite image synthesis~\cite{sastry2024geosynth,zhang2023diffucd,benidir2025change} rely on coarse-grained semantic supervision, typically using OpenStreetMap (OSM) raster tiles and/or global text prompts. Although OSM tiles are abundant and visually interpretable, they lack semantic depth. Distinct object classes, such as residential buildings, hospitals, or schools, are often depicted similarly in OSM stylings, failing to capture these semantic properties. In addition, image-level text supervision limits fine-grained control and does not allow more specific region-level editing.

To enable more expressive and semantically grounded image synthesis, we propose shifting from coarse, image-level conditioning to fine-grained, local-level representations. This shift enables users to specify distinct textual prompts for specific regions of an image, allowing for flexible editing, richer abstraction, and precise semantic control. However, achieving this level of generation requires a model that can accurately align textual descriptions with corresponding spatial regions, such as polygons or pixel masks. As illustrated in Figure~\ref{fig:task_description}, traditional image-level conditioning lacks fine-grained control, producing globally plausible imagery, but failing to reflect localized semantics. In contrast, our proposed approach enables fine-grained control by conditioning the synthesis process on sub-image annotations with detailed semantics.

OpenStreetMap (OSM), with its vast and growing repository of structured geographic annotations, is an ideal source of semantic grounding for spatial reasoning tasks. While recent vision-language models (VLMs) such as RemoteSAM~\cite{yao2025RemoteSAM} and RemoteCLIP~\cite{liu2024remoteclip} demonstrate pixel-level grounding capabilities, their vocabularies are typically constrained to general object categories (e.g., `car', `road') and lack alignment with the rich, structured, and domain-specific taxonomy used in OSM. These models are not trained to handle compositional or hierarchical tag structures (e.g., `building residential', `shop retail') that are common in OSM, and thus fall short in tasks that require detailed semantic understanding of geographic features. To address this gap, we propose learning fine-grained alignment between satellite imagery and OSM-style textual descriptions at the polygon level. Datasets like SkyScript~\cite{wang2024skyscript} offer global tag supervision, but do not include the vector geometries necessary for region-specific grounding. 

To this end, we propose a framework that enables local-level alignment between satellite imagery and OSM-based semantic descriptions. By grounding image synthesis at the polygon level, we allow for precise spatial control over generative models, enabling composition, editing, and abstraction beyond what is possible with coarse-level supervision.

\vspace{0.5em}

\noindent \textbf{Key Contributions:} We introduce a framework for pixel-level semantic control of satellite image synthesis. The contributions of our work are threefold:

\begin{enumerate}
    \item \textbf{COSA: Contrastive OSM-Satellite Alignment Vision-Language Model.} A model trained to align OSM tag descriptions and satellite imagery through polygon level contrastive learning. 
    \item \textbf{VectorSynth: Text-to-Image Generation with Pixel-Level Control.} A synthesis pipeline that enables compositional, fine-grained generation from multiple textual prompts, controlling content at the pixel-level.
    \item \textbf{OSM-Polygon Dataset.} A novel dataset that aligns satellite images with OSM polygon-level tags, allowing for fine-grained grounding of semantic regions.
\end{enumerate}
\section{Related Work}
\label{sec:related_work}

\textbf{Fine-grained Contrastive Learning.} Previous works have extended global vision-language models~\cite{radford2021learning, jia2021scaling, li2022blip} to capture token-level alignment of image and text for improved fine-grained understanding. RegionCLIP~\cite{zhong2022regionclip} uses pre-trained CLIP to label region-text pairs, guiding contrastive learning. LOUPE~\cite{li2022fine} generates semantic regions and performs region-text alignment. MaskCLIP~\cite{zhou2022maskclip} leverages CLIP's spatial tokens for dense prediction masks. Subsequent methods~\cite{lan2024clearclip, jing2024fineclip, zeng2024maskclip++} advance these approaches for tasks like open-vocabulary semantic segmentation.

However, these CLIP-based image encoders have inherently low latent resolutions that require upsampling for dense prediction tasks, limiting their capability in capturing fine-grained details~\cite{luddecke2022image}. In complex environments like urban remote sensing imagery, detailed information is lost when using these low-resolution latent models~\cite{kuckreja2024geochat,shabbirgeopixel}. FeatUp~\cite{fu2024featup} addresses this with a learnable feature upsampler. Applied to MaskCLIP, the model can capture more fine-grained text-image alignment.

While these approaches have better fine-grained text-image alignment, they keep the text encoder frozen to retain the benefits of CLIP pretraining. This poses limitations in the remote sensing domain, where textual semantics are often highly correlated. For example, `building height 5m' and `building height 30m' are linguistically similar, but may refer to visually distinct structures like a small house and a high-rise apartment, respectively. RemoteCLIP~\cite{liu2024remoteclip} attempts to address this by training both the image and text encoders for vision-language alignment in the remote sensing domain. Other approaches such as Sat2Cap~\cite{dhakal2024sat2cap} align satellite imagery with ground-level imagery to improve fine-grained understanding. However, all such models still have low-resolution latent image features. In this work, we propose a contrastive learning framework that jointly trains a high-resolution image encoder and a text encoder useful for fine-grained remote sensing synthesis.

\vspace{0.5em}

\noindent \textbf{Satellite Image Synthesis.} Recent advances in satellite image synthesis~\cite{khanna2023diffusionsat,yu2024metaearth} have shown promise across a range of remote sensing applications, including change detection~\cite{benidir2025change}, cloud removal~\cite{wang2023cloud}, and synthetic data generation for discriminative tasks~\cite{toker2024satsynth}. These approaches often rely on either modality-to-modality translation (e.g., SAR-to-optical~\cite{baisar2optical}) or style-conditioned generation using simplified semantic inputs. GeoSynth~\cite{sastry2024geosynth} is a notable work using semantic information for satellite image generation. It conditions a ControlNet~\cite{zhang2023adding} based diffusion model on OpenStreetMap (OSM) tile images that serve as a proxy for objects and land use structure. However, OSM stylings, while visually intuitive, are limited in both semantic depth and compositional control. They compress diverse geographic information into a fixed set of hand-crafted visual representations, which cannot easily capture multi-label, hierarchical, or region-specific semantics. Our work seeks to move beyond fixed visual stylings by leveraging the rich, structured tag data available in OSM. Rather than treating the OSM input as a 2D styled image, we encode raw OSM tag sets directly into the \emph{text space}, using vision-language models (VLMs) to establish semantically grounded and compositional controls. Our method opens up a new avenue for semantic satellite image synthesis by embedding structured geographic knowledge into a generative language-driven pipeline.
\section{Dataset}
\label{sec:dataset}

To achieve fine-grained conditioning, as seen in Figure~\ref{fig:task_description}, we construct a dataset coupling polygon-level OSM vector data with high-resolution satellite imagery. This enables fine-grained local alignment and compositional semantics for precise control over generated content.

To capture a wide range of urban layouts, we focus our data collection on five major cities. During training, we sample from Los Angeles, New York City, Paris, and Berlin. Each training city is split into spatial blocks and divided 60/20/20 into train/validation/test dataset splits. To assess the model’s ability to generalize beyond these known contexts, we designate Chicago as a holdout city to evaluate out-of-distribution performance. High-resolution satellite imagery is sourced from the Mapbox Static Tiles API\footnote{\url{https://docs.mapbox.com/api/maps/static-tiles/}} at zoom level 16, corresponding to a spatial resolution of approximately 0.6 meters per pixel. Each tile is 512×512 pixels, covering an area of roughly 300×300 meters, consistent with prior work~\cite{sastry2024geosynth} to support direct comparisons.

Vector data is obtained from GeoFabrik's OSM extracts~\cite{geofabrik_osm}. To retain only semantically relevant features visible in overhead imagery, we filter out point geometries, remove rare tags ($<$ 0.2\% of tiles), and keep only tiles with $\ge$70\% vector feature coverage. To further enhance structural diversity, we incorporate building height data from GlobalFootprintsLM~\cite{google_microsoft_open_buildings}, enriching the representation of vertical variation across scenes. 

\begin{figure}
  \centering
  \includegraphics[width=0.85\linewidth]{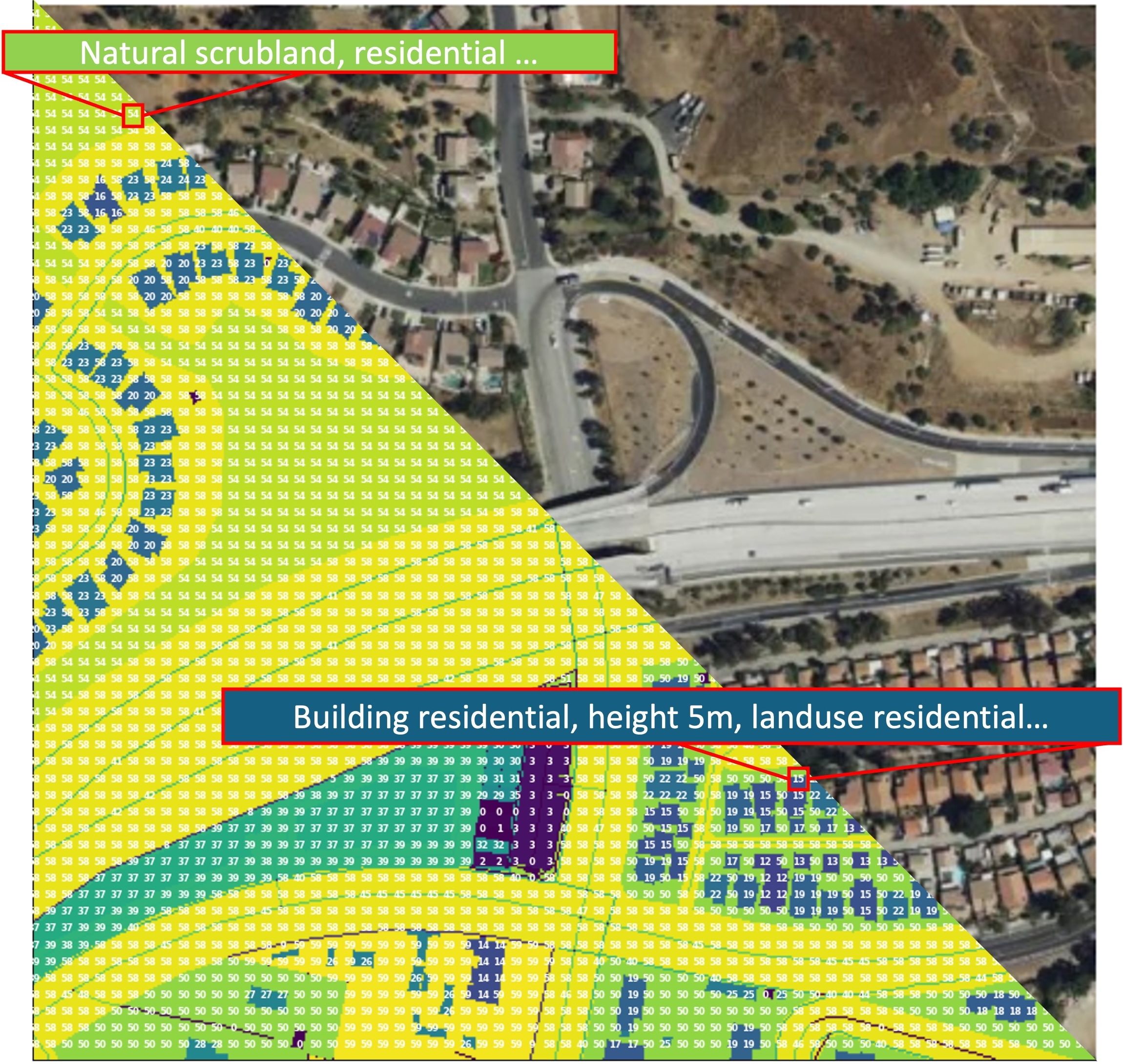}
  \caption{Illustration of pixel-level tag assignments. Each pixel inherits tags from overlapping polygons, resulting in compositional tag lists used for downstream learning tasks.}
  \label{fig:pixel_tags}
\end{figure}

To represent semantic content at the pixel level, as seen in Figure~\ref{fig:pixel_tags}, we render the filtered vector features by collecting tags from overlapping annotations per pixel. Each pixel gets a multi-tag composition, and nearby pixels with identical compositions form polygon instances. For example, a single pixel might inherit tags such as [`building residential', `place island', `height 6m'], reflecting multiple overlapping semantic layers.

We illustrate the richness and granularity of the tag annotations by visualizing multi-tag composition overlays on a sample tile shown in Figure~\ref{fig:pixel_tags}. The overlay reveals dense and semantically consistent tagging across spatial structures, such as roads, residential blocks, and natural features, highlighting the high quality and compositional expressiveness of our annotations.

We also render all tiles using custom Mapbox styles to generate stylized OSM maps to conduct consistent evaluation against previous work~\cite{sastry2024geosynth}. Finally, we caption each satellite image using LLaVA~\cite{liu2024visual}, a multimodal vision language model. The prompt used for captioning is: `Describe the contents of the image'. The final OSM-Satellite dataset includes approximately 1,000 unique OSM tags and over 400,000 unique tag combinations associated with individual pixels. We generate around 20,000 image tiles, each paired with satellite imagery, polygon-level vector annotations, pixel-level semantic masks, global satellite descriptions and corresponding OSM stylizations. This dataset serves as the foundation for pixel-level contrastive learning and image synthesis tasks.


\section{Methodology}
\label{sec:methodology}

In this section, we describe our proposed approach to build a semantically aligned pixel-level vision-language model and a fine-grained satellite image synthesis framework.
\subsection{Polygon-Level Contrastive Learning}

\begin{figure}
    \centering
    \includegraphics[width=0.95\columnwidth]{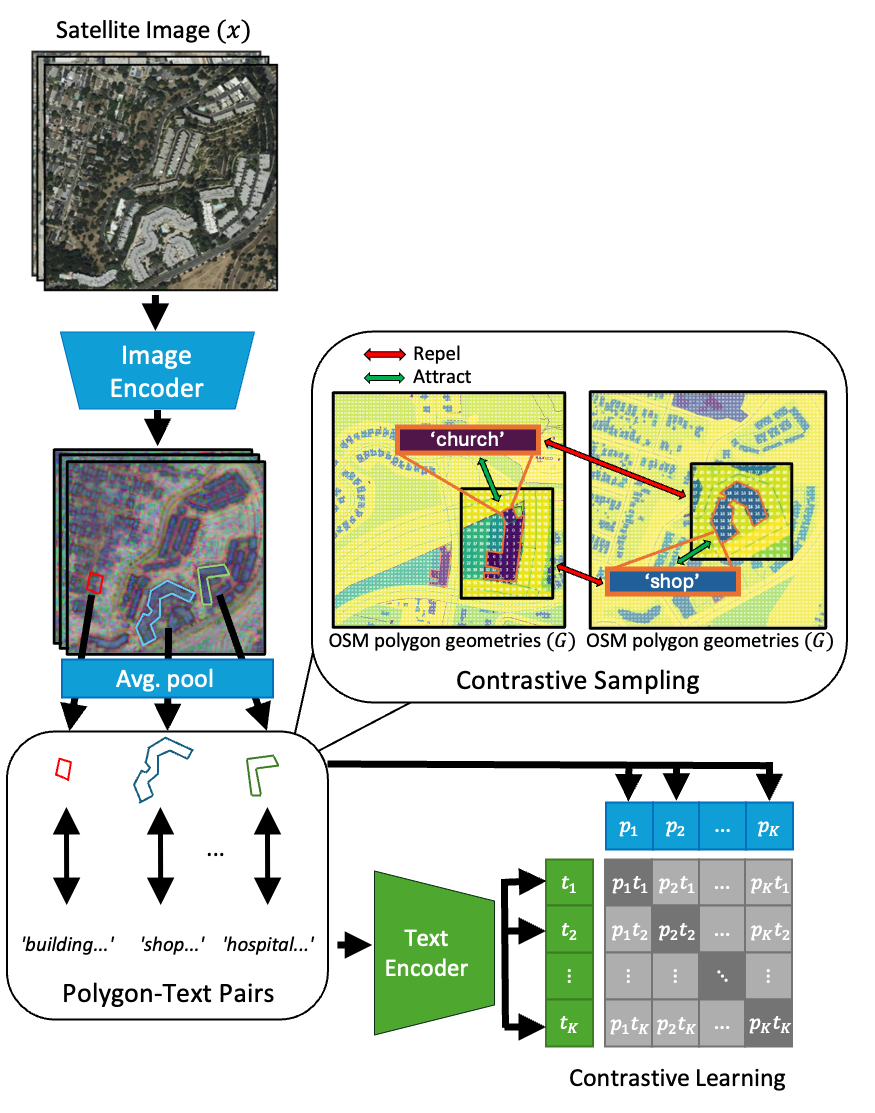}
    \caption{Architecture overview showing dual encoders for satellite imagery and OSM tag descriptions, with polygon-guided average pooling to extract region-specific embeddings. We align polygon embeddings with grounded OSM tags, enabling fine-grained spatial conditioning for satellite image synthesis.}
    \label{fig:OSM_CLIP}
\end{figure}

We learn a textual embedding space for OSM tags aligned with satellite imagery to enable fine-grained conditioning during synthesis. As described in Section~\ref{sec:dataset}, polygons define spatial units associated with multi-tag compositions in the image. We apply contrastive learning to pull closer the embeddings of aligned polygon--tag composition pairs (e.g., a group of satellite pixels and its corresponding multi-tag composition), while pushing apart the embeddings of dissimilar pairs (e.g., polygons associated with different multi-tag compositions), as seen in Figure~\ref{fig:OSM_CLIP}.

Previous satellite image-text contrastive learning frameworks~\cite{liu2024remoteclip} aim to enhance image encoders, so their representations more closely align with a pretrained text embedding space, such as CLIP. These models are often optimized for image retrieval or segmentation, but not text-guided generation. In contrast, we focus on improving the text encoder to align better with the image embedding space for the purpose of fine-grained generation. 

To support this, we opt for dense pixel-level representations. Previous dense contrastive frameworks form contrastive pairs at the patch level~\cite{zhou2022maskclip} or use self-supervised spatial cues~\cite{chen2023revisiting, yao2021filip}, while our approach leverages vector polygon annotations with explicit tag labels from OSM. These polygons are directly aligned with image regions, which enables polygon-guided contrastive learning.

\vspace{0.5em}
\noindent \textbf{Architecture.} Our model, \textbf{COSA}, is shown in Figure~\ref{fig:OSM_CLIP}. The architecture consists of a learnable image encoder $f_{\text{img}}$, a learnable text encoder $f_{\text{text}}$, and a polygon-guided contrastive loss objective that aligns OSM tag compositions with corresponding polygon image features. 

Let $x \in \mathbb{R}^{3 \times H \times W}$ denote a satellite image with height $H$ and width $W$. Let $\mathcal{C} = \{c_1, \dots, c_K\}$ be OSM multi-tag compositions corresponding to polygon geometries $G = \{g_1, \dots, g_K\}$ in the image, where each multi-tag composition $c_i$ is a sentence. The text and image encoders process $c_i$ and $x$ respectively to produce a corresponding text embedding $e$ and dense image embeddings $z_{\text{img}} \in \mathbb{R}^{D \times H' \times W'}$, where $D$ is the embedding dimension and $(H', W')$ is the spatial resolution of the image feature map.

\begin{figure*}[t!]
  \centering
  \includegraphics[width=\linewidth]{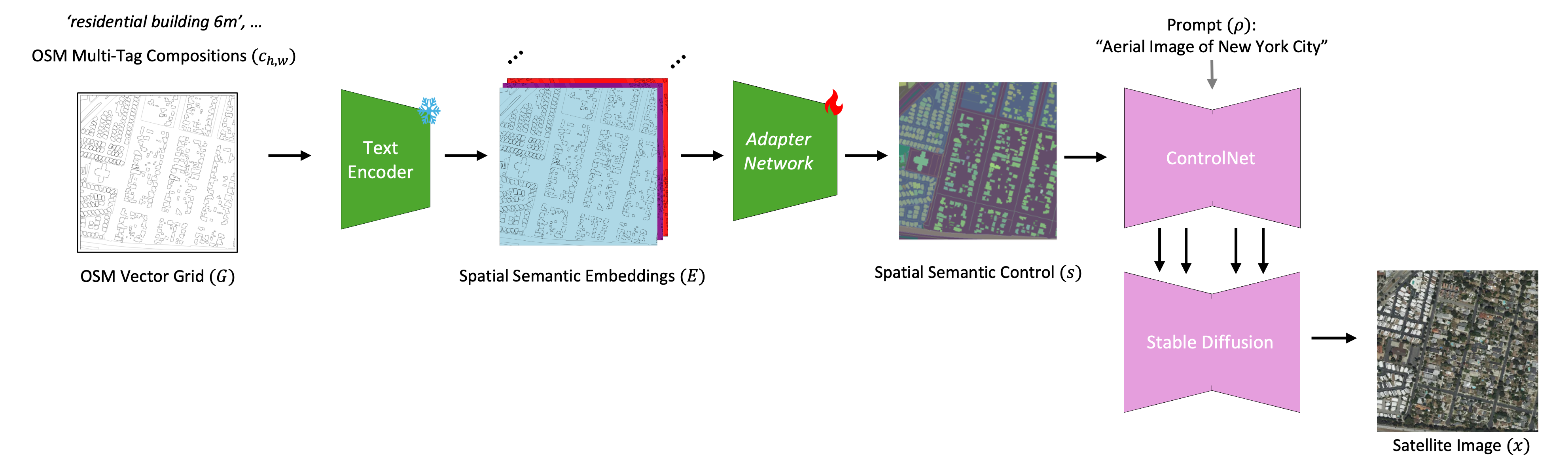}
  \caption{
    This figure presents our semantic-guided image synthesis pipeline which employs a pretrained text encoder to generate dense pixel-level control from input vector geometry.
  }
  \label{fig:semantic_guided_synthesis}
\end{figure*}
\vspace{0.5em}
\noindent \textbf{Polygon-Guided Contrastive Loss.}
We aim to generate polygon-text contrastive pairs for training. For each polygon geometry $g_i$ in the satellite image $x$, let $M_{g_i} \in \{0, 1\}^{H' \times W'}$ be a binary mask indicating the spatial extent of the polygon on the image feature map. We obtain this binary mask by interpolating and thresholding from the resolution of the image features. The polygon embedding $p_i \in \mathbb{R}^D$ is computed by average pooling $z_{\text{img}}$ over the masked region: 
\begin{equation}
p_i = \frac{1}{\sum{M_{g_i}(h, w)}} \sum_{h=1}^{H'} \sum_{w=1}^{W'} M_{g_i}(h, w) \cdot z_{\text{img}}[:, h, w]
\end{equation}
Given $K$ polygon-text pairs $\{(p_i, e_i)\}_{i=1}^{K}$, we use a symmetric InfoNCE loss which is defined as follows: 
\begin{equation}
\begin{aligned}
\mathcal{L}_{p, e} = 
- \frac{1}{2K} \sum_{i=1}^{K} \Bigg[
    & \log \frac{
        \exp\left( \text{sim}(p_i, e_i) / \tau \right)
    }{
        \sum_{j=1}^{K} \exp\left( \text{sim}(p_i, e_j) / \tau \right)
    } \\
    +\; & \log \frac{
        \exp\left( \text{sim}(e_i, p_i) / \tau \right)
    }{
        \sum_{j=1}^{K} \exp\left( \text{sim}(e_i, p_j) / \tau \right)
    }
\Bigg]
\end{aligned}
\end{equation}
where $\text{sim}(\cdot, \cdot)$ denotes cosine similarity and $\tau$ is a learnable temperature parameter.

\subsection{Image Synthesis}

\noindent \textbf{Architecture.}
We train conditional generative models to synthesize satellite images $\mathit{x}$ given an image-level natural language text description $\rho$ and spatial semantic control $\mathit{s}$ derived from OpenStreetMap (OSM). Specifically, we optimize a latent diffusion model to approximate the conditional distribution $p(\mathit{x} \mid \rho, \mathit{s})$.


We build upon the ControlNet~\cite{zhang2023adding} architecture, which extends pre-trained diffusion models by incorporating additional control inputs. ControlNet consists of a trainable copy of the encoding layers of the base diffusion model, connected via zero-initialized convolution layers. This design preserves the original model's capabilities while enabling additional control. The control branch processes the conditioning information at multiple scales and feeds it into the main U-Net through residual connections. 

Figure~\ref{fig:semantic_guided_synthesis} illustrates our complete pipeline. The generation is guided by two controls: a global text prompt $\rho$ and a spatial semantic control $\mathit{s}$. 

We derive $\mathit{s}$ from OSM vector geometries. First, we render the OSM data into a grid $\mathit{G} \in \mathcal{C}^{H \times W}$, where each pixel $(h, w)$ contains a multi-tag composition $\mathit{c}_{h, w} \in \mathcal{C}$. Next, we encode each composition with a text encoder $\mathit{T}$, producing spatial semantic embeddings $\mathit{E} \in \mathbb{R}^{D \times H \times W}$:
\begin{equation}
\mathit{E}[h, w] = \mathit{T}(\mathit{c}_{h, w})
\end{equation}
To align $\mathit{E}$ with ControlNet, we pass it through a lightweight adapter $\mathcal{A}$, yielding a 3-channel raster:
\begin{equation}
\mathit{s} = \mathcal{A}(\mathit{E}) \in \mathbb{R}^{3 \times H \times W}
\end{equation}
The spatial semantic control $\mathit{s}$ provides both layout guidance and fine-grained OSM tag information, while the text prompt $\rho$ provides global context.

Finally, ControlNet conditions the diffusion process on both $\mathit{s}$ and $\rho$, training the denoising network $\epsilon_\theta$ according to the objective:
\begin{equation}
\mathcal{L} = \mathbb{E}_{\mathit{z}_0, \mathit{s}, \rho, \epsilon} \left[ \|\epsilon - \epsilon_\theta(\mathit{z}_t, \mathit{s}, \rho)\|_2^2 \right]
\end{equation}
where $\mathit{z}_t$ is the noisy latent representation at diffusion timestep $t$.

\noindent During inference, users render OSM multi-tag compositions into a pixel grid, which is encoded into the same control representation $\mathit{s}$. ControlNet then synthesizes images consistent with these pixel-wise semantics.
\begin{figure*}[!ht]
\centering
\includegraphics[width=\textwidth]{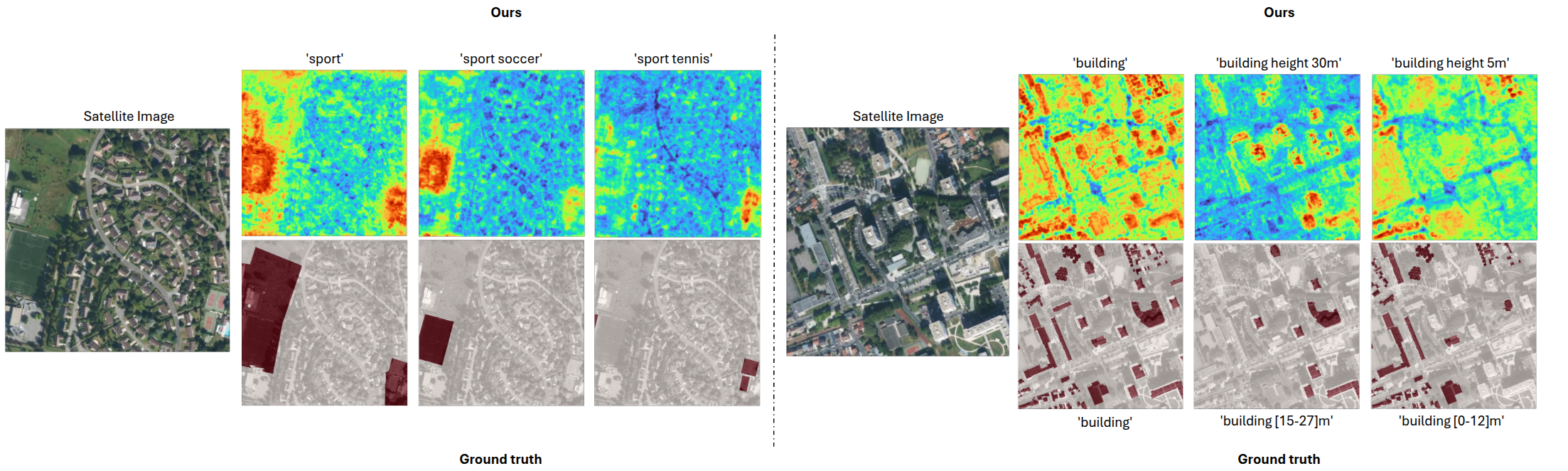}
\caption{Similarity heatmaps given different text queries highlighting the fine-grained understanding of the proposed contrastive training. The bottom row shows the respective ground truth polygons. The alignment shows the ability of our approach to disentangle correlated semantic structures in OSM tag text.}
\label{fig:sports_and_building_heatmap}
\end{figure*}

\subsection{Implementation}
For the COSA VLM, we use a SatlasNet~\cite{bastani2023satlaspretrain} backbone with a learnable MLP adapter network as an image encoder. We use CLIP as our default learnable text encoder, but also experiment with BERT~\cite{devlin2019bert} and E5~\cite{wang2022text}, which are strong sentence level embedding models.  For the image synthesis framework, we use Stable Diffusion v2.1~\cite{rombach2022high}. 

We precompute text embeddings of all OSM taglists in our dataset for training- and inference-time efficiency. We train six model variants, experimenting with different text encoders and adapter network architectures. For the adapter network, we experimented with deeper convolutional stacks and residual connections, but found that a 2D convolution followed by a sigmoid activation consistently yielded the best results. Following ControlNet~\cite{zhang2023adding}, we apply random prompt masking. This encourages the model to leverage spatial semantics from the control image even when text is absent, improving robustness and generalization. Each model was trained on a single NVIDIA H100 GPU (80GB) for a total of 24 hours, using the Adam optimizer with a learning rate of $1 e{-5}$ and a batch size of 8. 
\section{Results and Discussion}
\label{sec:results_discussion}

We conduct extensive evaluations of VectorSynth, with ablations to assess contributions.

\subsection{Cross-Modal Evaluation}

We evaluate COSA, our contrastively trained vision-language model, on its ability to align polygon-level satellite imagery with OSM multi-tag composition. Performance is assessed through cross-modal retrieval and polygon-level tag prediction, testing fine-grained semantic grounding and generalization.

\begin{table}[t]
\centering
\small
\resizebox{\linewidth}{!}{
\begin{tabular}{lcccc}
\toprule
\textbf{VLM} & \textbf{B@1 ↑} & \textbf{Sem@20 ↑} & \textbf{R@5 ↑} & \textbf{R@10 ↑} \\
\midrule
CLIP        & $88.68$ & $88.30$ & $25.48$ & $46.79$ \\
RemoteCLIP  & $87.83$ & $87.39$ & $26.58$ & $51.73$ \\
\midrule
Ours (E5)        & $89.86$ & $90.37$ & $21.14$ & $48.14$ \\
Ours (BERT-base) & \underline{90.11} & \underline{91.64} & \underline{29.49} & \underline{53.10} \\
Ours (CLIP)      & \textbf{91.61} & \textbf{92.47} & \textbf{32.40} & \textbf{54.06} \\
\bottomrule
\end{tabular}
}
\vspace{0.5em}
\caption{
Polygon-to-text retrieval results. We compare baseline VLMs with our COSA variants that use different text encoders. \textbf{Note:} B@1 is BERTScore@1, Sem@20 is Semantic-nDCG@20, R@K is Recall@K. Best results are \textbf{bold} and second best are \underline{underlined}.
}
\label{tab:retrieval_eval}
\end{table}

\begin{table}[t]
\centering
\small
\resizebox{\linewidth}{!}{
\begin{tabular}{lccc}
\toprule
\textbf{VLM} & \textbf{Parent Acc. ↑} & \textbf{Child Acc. ↑} & \textbf{Mixed F1 ↑} \\
\midrule
CLIP        & $43.16$ & $29.41$ & $0.171$ \\
RemoteCLIP  & $44.34$ & $31.58$ & $0.172$ \\
Ours        & \textbf{82.84} & \textbf{77.09} & \textbf{0.272} \\
\bottomrule
\end{tabular}
}
\vspace{0.5em}
\caption{
OSM tag prediction results. `Ours' denotes our COSA VLM with a CLIP text encoder. 
\textbf{Note:} Parent/Child accuracy measure correctness at broad vs.\ fine-grained tag levels, and Mixed F1 is their averaged F1. 
Best results are \textbf{bold}.
}
\label{tab:tag_pred_eval}
\end{table}

\vspace{0.5em}

\begin{table*}[t!]
\centering
\small
\begin{tabular}{@{}llcccccc@{}}
\toprule
\multirow{2}{*}{\textbf{Model}} & \multirow{2}{*}{\textbf{\begin{tabular}{@{}c@{}}Finetuned \\ Text Encoder\end{tabular}}} &
\multicolumn{3}{c}{\textbf{In-Distribution Test}} & 
\multicolumn{3}{c}{\textbf{Out-of-Distribution Test}} \\
\cmidrule(lr){3-5} \cmidrule(l){6-8}
 &  & \textbf{FID} $\downarrow$ & \textbf{SSIM} $\uparrow$ & \textbf{PSNR} $\uparrow$ & \textbf{FID} $\downarrow$ & \textbf{SSIM} $\uparrow$ & \textbf{PSNR} $\uparrow$ \\ 
\midrule
GeoSynth-OSM~\cite{sastry2024geosynth} & \xmark & 95.30 & 0.16 & 9.92 & 108.33 & 0.15 & 9.89 \\
\midrule
\multicolumn{8}{c}{\textbf{Ours (VectorSynth)}} \\
\midrule
BERT        & \xmark & 52.16 & 0.15 & 12.59 & 72.15 & 0.10 & 10.91 \\
E5          & \xmark & 46.71 & 0.15 & 12.92 & 63.66 & 0.12 & 11.24 \\
CLIP        & \xmark & 33.07 & 0.20 & 14.04 & 45.13 & 0.16 & 12.03 \\
RemoteCLIP  & \xmark & 48.75 & 0.14 & 12.89 & 61.54 & 0.11 & 11.58 \\
\midrule
\multicolumn{8}{c}{\textbf{Ours (VectorSynth + COSA)}} \\
\midrule
COSA-BERT & \cmark & 40.61 & 0.15 & 12.66 & 67.72 & 0.11 & 11.07 \\
COSA-E5   & \cmark & 57.06 & 0.17 & 13.27 & 68.22 & 0.14 & 11.69 \\
COSA-CLIP & \cmark & \textbf{29.20} & \textbf{0.21} & \textbf{14.15} & \textbf{41.12} & \textbf{0.17} & \textbf{12.10} \\
\midrule
Gain (\%) &  & \textcolor{Green}{+69.36} & \textcolor{Green}{+31.25} & \textcolor{Green}{+42.64} & \textcolor{Green}{+62.04} & \textcolor{Green}{+13.33} & \textcolor{Green}{+22.35} \\
\bottomrule
\end{tabular}
\caption{Quantitative evaluation of different text encoders used in control generation. We show VectorSynth with and without contrastive OSM-Satellite alignment on in-distribution and out-of-distribution test sets used for tag conditions at the sub-image level. \textbf{Note}: All models have the same global text encoder for the SD2.1 base model.} 
\label{tab:quant_comparison}
\end{table*}

\noindent \textbf{Cross-Modal Retrieval.} We report BERTScore@1~\cite{zhang2019bertscore} for top-1 semantic similarity, semantic nDCG@20~\cite{jarvelin2002cumulated} for ranked semantic relevance, and Recall@5/10 for retrieval accuracy. Table~\ref{tab:retrieval_eval} reports polygon-to-text retrieval results on the test set. The VLM baselines, CLIP and RemoteCLIP, achieve reasonable performance, however our COSA model variants show consistent improvements across all metrics, with the CLIP-based text encoder achieving the highest BERTScore@1 (91.61), semantic nDCG@20 (92.47), Recall@5 (32.40), and Recall@10 (54.06). These gains highlight the importance of the choice of the text encoder. Furthermore, our contrastive training provides better alignment for fine-grained semantic retrieval compared to baseline VLMs. 

\vspace{0.5em}

\noindent \textbf{Polygon-Level Tag Prediction.} Beyond retrieval, we also evaluate our model's ability to directly predict OSM tags for polygons. We evaluate this task using three metrics: Parent accuracy, which measures correctness for parent tags (i.e. `building') to capture broader semantic categories; Child accuracy, which measures correctness for child tags within a parent (i.e. `apartments') to capture more fine-grained categories; Mixed F1 score, the average of parent- and child-level F1. Compared to CLIP and RemoteCLIP, COSA substantially improves across parent- and child-level accuracy, along with Mixed F1 score. This demonstrates that contrastive training not only enhances retrieval but also enables stronger tag prediction, opening the door to effective pseudo-labeling for sparsely annotated OSM regions. 




We visualize the normalized cosine similarity matrix of tag embeddings in Figure~\ref{fig:sports_and_building_heatmap} to illustrate the model’s fine-grained semantic understanding. After contrastive training, embeddings capture more structured and meaningful relationships, particularly among closely related categories (e.g., `sport tennis' vs. `sport soccer'). These patterns reflect improved sensitivity to subtle distinctions in OSM tags, such as different types of sports facilities and buildings.

\begin{figure}
    \centering
    \includegraphics[width=0.95\columnwidth]{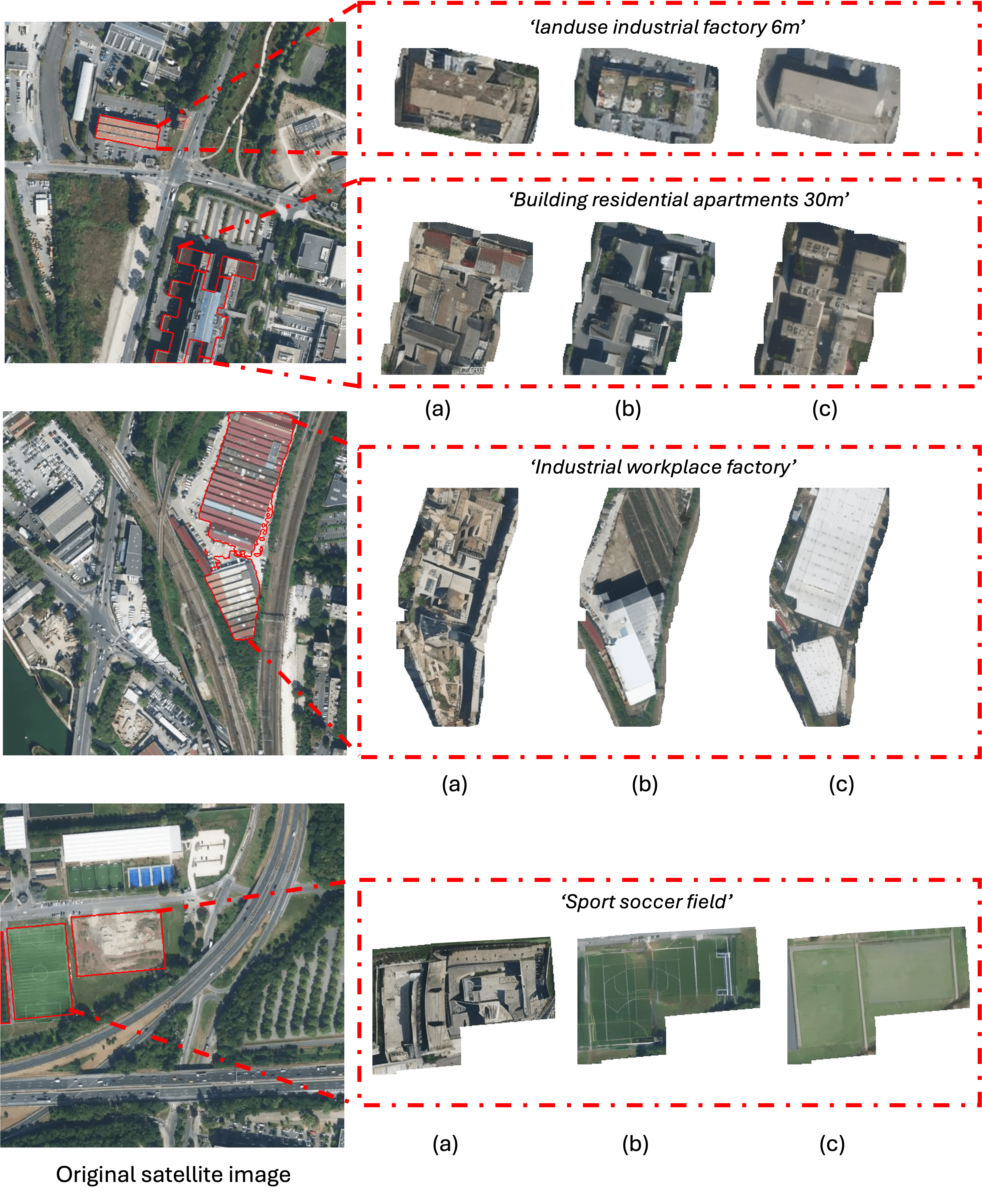}
    \caption{Comparison of fine-grained semantic edits. Each set shows the local caption used for editing across models: (a) GeoSynth, (b) GeoSynth w/ Inpainting and (c) VectorSynth}
    \label{fig:semantic_edits}
\end{figure}

\begin{figure*}[t!]
  \centering
  \includegraphics[width=\linewidth]{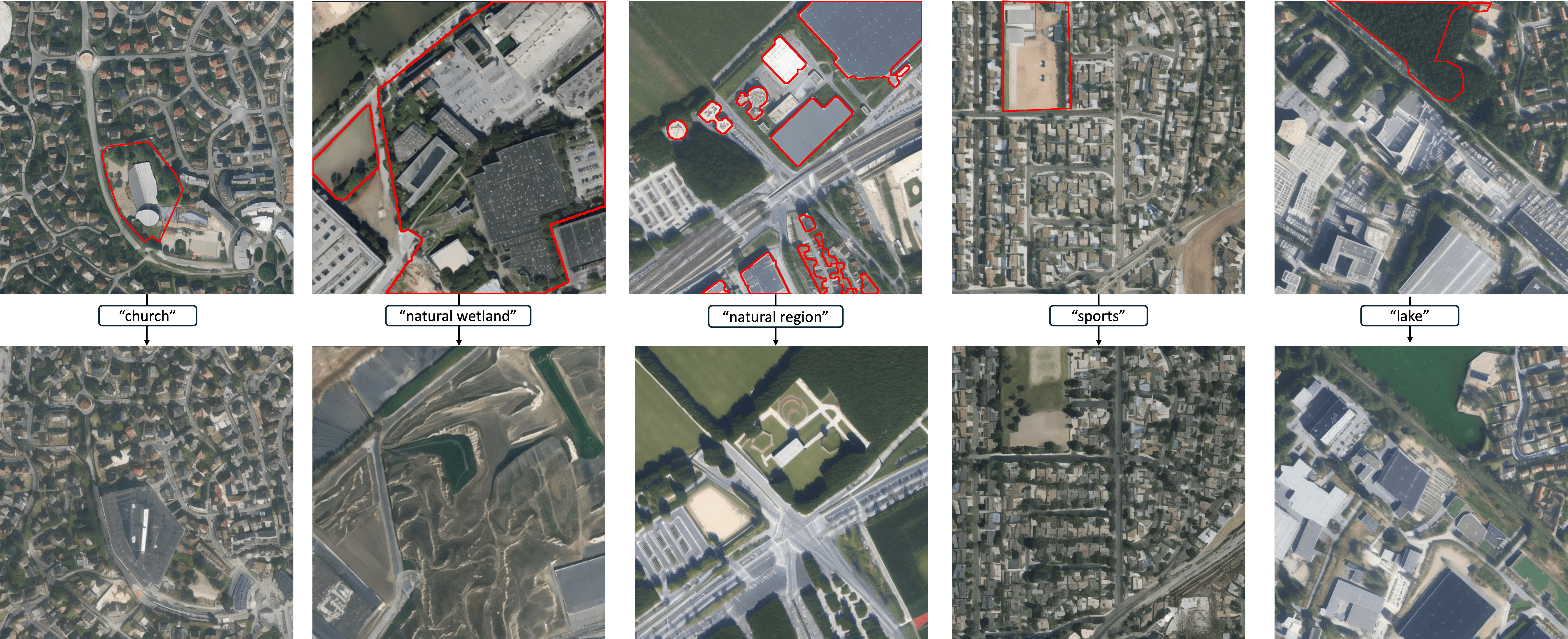}
  \caption{
    Examples of semantic edits across urban planning, and landuse generation applications.
  }
  \label{fig:vecsynth_examples}
\end{figure*}

\subsection{Semantically Grounded Image Synthesis}

We evaluate the quality and controllability of satellite image synthesis under various input representations and conditioning types. Specifically, we assess: (1) the impact of different embedding sources for pixel-level control, and (2) fine-grained semantic editing capabilities.

\vspace{0.5em}

\noindent \textbf{Embedding Sources for Control.} We evaluate three different approaches to synthesize satellite imagery conditioned on different representations of OpenStreetMap (OSM) control signals. The first framework, \textbf{GeoSynth-OSM}, follows the Geosynth-OSM baseline~\cite{sastry2024geosynth} and uses rasterized OSM tiles as direct pixel-level input. The second framework, \textbf{VectorSynth},  incorporates OSM tags as textual input by embedding them with a pretrained text encoder (e.g., CLIP), then projecting the resulting embeddings to a pixel-level control map. The third framework, \textbf{VectorSynth-COSA}, builds on this approach but replaces the off-the-shelf text encoder with our aforementioned COSA model's text encoder. Across both text-based variants, we experiment with different text encoders and study the effect of contrastive learning on grounding semantic control in image synthesis. All results reported use a 2D convolutional adapter network.

Table~\ref{tab:quant_comparison} highlights the quantitative gains of our approach, showing strong and consistent improvements across all standard metrics. We evaluate the quality of generated images using three standard metrics: Fréchet Inception Distance (FID)~\cite{heusel2017gans} to measure distributional similarity to real images, Structural Similarity Index (SSIM)~\cite{wang2004imagessim} to assess perceptual quality, and Peak Signal-to-Noise Ratio (PSNR) to quantify pixel-level reconstruction accuracy. Our method significantly outperforms GeoSynth-OSM when using text-based control inputs. We see further gains when using COSA-aligned encoders over their vanilla (non-aligned) counterparts. Notably, we also compare our method with RemoteCLIP's text encoder specifically tuned for remote sensing tasks and see that our COSA-aligned text encoders consistently perform better, demonstrating stronger semantic grounding.

\begin{table}[t]
\renewcommand{\arraystretch}{0.95}
\setlength{\tabcolsep}{5pt}
\centering
\small
\begin{tabular}{lcc}
\toprule
\textbf{Category} & \textbf{GeoSynth + Inpaint} & \textbf{VectorSynth} \\
\midrule
building & 18.98 & 13.76 \\
natural  & 29.50 & 28.40 \\
place    & 26.32 & 21.30 \\
landuse  & 17.21 & 12.95 \\
highway  & 23.28 & 16.29 \\
\midrule
Combined  & 16.19 & \textbf{11.34} \\
\bottomrule
\end{tabular}
\caption{
FID scores ($\downarrow$) for semantic edits across categories. GeoSynth + Inpaint uses Stable Diffusion inpainting~\cite{rombach2022high} with GeoSynth weights, while VectorSynth uses our standard pipeline. \textbf{Note:} For fairness, non-edited regions are preserved from the original image, lowering FID relative to full image synthesis.
}
\label{tab:edit_results}
\end{table}

\vspace{0.5em}

\noindent \textbf{Fine-Grained Semantic Control.} To evaluate semantic editing capabilities, we introduce targeted edits and examine their localized impact to the output image. Figure~\ref{fig:semantic_edits} shows diverse examples across semantic categories and spatial contexts, demonstrating the model’s ability to produce precise, semantically meaningful edits. We compare VectorSynth to GeoSynth~\cite{sastry2024geosynth}, and Stable Diffusion Inpainting~\cite{rombach2022high} using GeoSynth weights. VectorSynth produces more realistic and coherent edits: buildings exhibit consistent structure and alignment, while features such as soccer fields appear more regular and well-formed. Quantitatively, we evaluate semantic edits in Table~\ref{tab:edit_results}. VectorSynth editing achieves lower FID scores across various semantic categories, indicating high fidelity and better alignment with the original distribution. When all categories are combined, VectorSynth also outperforms. Note that in the combined FID, all edits are aggregated into a larger, lower-variance set, which yields lower scores than per-category metrics. Additional capabilities are also shown in Figure~\ref{fig:vecsynth_examples}, lending to potential applications of this model such as in urban planning and landuse generation. 

\section{Conclusion}
\label{sec:conclusion}

We introduced VectorSynth, a novel approach for satellite image synthesis that provides fine-grained pixel-level semantic control, moving beyond the coarse-grained conditioning of prior work, which relied on broad categories, such as buildings, parks, and roads. By allowing users to define more detailed semantics for different regions in the image, VectorSynth enables a broad range of applications, from data generation for machine learning models to citizen-driven urban design. Our approach aligns the representation space using polygon-level contrastive learning, outperforming strong, off-the-shelf embedding networks. The current design is well-suited to editing scenarios, where precise, localized control is essential. Future work includes enhancing the model's ability to learn from sparse or incomplete annotations, thereby increasing its applicability in data-limited settings. Additionally, enabling the network to hallucinate uncontrolled map regions more effectively would allow users to specify fewer semantic regions while still generating coherent, high-quality scenes.
\clearpage
\section*{Acknowledgments}
\label{sec:acknowledgments}
This research was supported by the Applied Research Laboratory for Intelligence and Security (ARLIS) at the University of Maryland, a University Affiliated Research Center of the Office of the Undersecretary of Defense for Intelligence and Security.
This research also used the TGI RAILs advanced compute and data resource, which is supported by the National Science Foundation (award OAC-2232860) and the Taylor Geospatial Institute. The code for this project is highly inspired by Sastry et al.~\cite{sastry2024geosynth} GitHub repository: \url{https://github.com/mvrl/GeoSynth}.
{
    \small
    \bibliographystyle{ieeenat_fullname}
    \bibliography{main}

@String(CVPR= {IEEE Conf. Comput. Vis. Pattern Recog.})

@String(ECCV= {Eur. Conf. Comput. Vis.})

@String(AAAI = {AAAI})

@String(CVPR  = {CVPR})

@String(ECCV  = {ECCV})

@ARTICLE{haklay2008osm,
  author={Haklay, Mordechai and Weber, Patrick},
  journal={IEEE Pervasive Computing}, 
  title={OpenStreetMap: User-Generated Street Maps}, 
  year={2008},
  volume={7},
  number={4},
  pages={12-18},
  doi={10.1109/MPRV.2008.80}}

@article{ramesh2022hierarchical,
  title={Hierarchical text-conditional image generation with clip latents},
  author={Ramesh, Aditya and Dhariwal, Prafulla and Nichol, Alex and Chu, Casey and Chen, Mark},
  journal={arXiv preprint arXiv:2204.06125},
  volume={1},
  number={2},
  pages={3},
  year={2022}
}

@inproceedings{rombach2022high,
  title={High-resolution image synthesis with latent diffusion models},
  author={Rombach, Robin and Blattmann, Andreas and Lorenz, Dominik and Esser, Patrick and Ommer, Bj{\"o}rn},
  booktitle={Proceedings of the IEEE/CVF conference on computer vision and pattern recognition},
  pages={10684--10695},
  year={2022}
}

@inproceedings{kawar2023imagic,
  title={Imagic: Text-based real image editing with diffusion models},
  author={Kawar, Bahjat and Zada, Shiran and Lang, Oran and Tov, Omer and Chang, Huiwen and Dekel, Tali and Mosseri, Inbar and Irani, Michal},
  booktitle={Proceedings of the IEEE/CVF Conference on Computer Vision and Pattern Recognition},
  pages={6007--6017},
  year={2023}
}

@inproceedings{brooks2023instructpix2pix,
  title={Instructpix2pix: Learning to follow image editing instructions},
  author={Brooks, Tim and Holynski, Aleksander and Efros, Alexei A},
  booktitle={Proceedings of the IEEE/CVF Conference on Computer Vision and Pattern Recognition},
  pages={18392--18402},
  year={2023}
}

@inproceedings{luddecke2022image,
  title={Image segmentation using text and image prompts},
  author={L{\"u}ddecke, Timo and Ecker, Alexander},
  booktitle={Proceedings of the IEEE/CVF conference on computer vision and pattern recognition},
  pages={7086--7096},
  year={2022}
}

@inproceedings{dhakal2024sat2cap,
  title={Sat2cap: Mapping fine-grained textual descriptions from satellite images},
  author={Dhakal, Aayush and Ahmad, Adeel and Khanal, Subash and Sastry, Srikumar and Kerner, Hannah and Jacobs, Nathan},
  booktitle={Proceedings of the IEEE/CVF Conference on Computer Vision and Pattern Recognition},
  pages={533--542},
  year={2024}
}

@inproceedings{kuckreja2024geochat,
  title={Geochat: Grounded large vision-language model for remote sensing},
  author={Kuckreja, Kartik and Danish, Muhammad Sohail and Naseer, Muzammal and Das, Abhijit and Khan, Salman and Khan, Fahad Shahbaz},
  booktitle={Proceedings of the IEEE/CVF Conference on Computer Vision and Pattern Recognition},
  pages={27831--27840},
  year={2024}
}

@inproceedings{shabbirgeopixel,
  title={GeoPixel: Pixel Grounding Large Multimodal Model in Remote Sensing},
  author={Shabbir, Akashah and Zumri, Mohammed and Bennamoun, Mohammed and Khan, Fahad Shahbaz and Khan, Salman},
  booktitle={Forty-second International Conference on Machine Learning},
year={2025}

}

@article{fu2024featup,
  title={FeatUp: A model-agnostic framework for features at any resolution},
  author={Fu, Stephanie and Hamilton, Mark and Brandt, Laura and Feldman, Axel and Zhang, Zhoutong and Freeman, William T},
  journal={arXiv preprint arXiv:2403.10516},
  year={2024}
}

@inproceedings{devlin2019bert,
  title={Bert: Pre-training of deep bidirectional transformers for language understanding},
  author={Devlin, Jacob and Chang, Ming-Wei and Lee, Kenton and Toutanova, Kristina},
  booktitle={Proceedings of the 2019 conference of the North American chapter of the association for computational linguistics: human language technologies, volume 1 (long and short papers)},
  pages={4171--4186},
  year={2019}
}

@article{wang2022text,
  title={Text embeddings by weakly-supervised contrastive pre-training},
  author={Wang, Liang and Yang, Nan and Huang, Xiaolong and Jiao, Binxing and Yang, Linjun and Jiang, Daxin and Majumder, Rangan and Wei, Furu},
  journal={arXiv preprint arXiv:2212.03533},
  year={2022}
}

@inproceedings{bastani2023satlaspretrain,
  title={SatlasPretrain: A large-scale dataset for remote sensing image understanding},
  author={Bastani, Favyen and Wolters, Piper and Gupta, Ritwik and Ferdinando, Joe and Kembhavi, Aniruddha},
  booktitle={Proceedings of the IEEE/CVF International Conference on Computer Vision},
  pages={16772--16782},
  year={2023}
}

@inproceedings{toker2024satsynth,
  title={Satsynth: Augmenting image-mask pairs through diffusion models for aerial semantic segmentation},
  author={Toker, Aysim and Eisenberger, Marvin and Cremers, Daniel and Leal-Taix{\'e}, Laura},
  booktitle={Proceedings of the IEEE/CVF Conference on Computer Vision and Pattern Recognition},
  pages={27695--27705},
  year={2024}
}

@inproceedings{wang2024skyscript,
  title={SkyScript: A large and semantically diverse vision-language dataset for remote sensing},
  author={Wang, Zhecheng and Prabha, Rajanie and Huang, Tianyuan and Wu, Jiajun and Rajagopal, Ram},
  booktitle={Proceedings of the AAAI Conference on Artificial Intelligence},
  volume={38},
  number={6},
  pages={5805--5813},
  year={2024}
}

@article{li2023gligen,
  title={GLIGEN: Open-Set Grounded Text-to-Image Generation},
  author={Li, Yuheng and Liu, Haotian and Wu, Qingyang and Mu, Fangzhou and Yang, Jianwei and Gao, Jianfeng and Li, Chunyuan and Lee, Yong Jae},
  journal={CVPR},
  year={2023}
}

@misc{geofabrik_osm,
  author       = {{Geofabrik GmbH}},
  title        = {Geofabrik Download Server},
  year         = {2024},
  url          = {https://download.geofabrik.de/},
  note         = {Accessed: 2025-07-17},
  howpublished = {\url{https://download.geofabrik.de/}}
}

@InProceedings{zhou2022maskclip,
    author = {Zhou, Chong and Loy, Chen Change and Dai, Bo},
    title = {Extract Free Dense Labels from CLIP},
    booktitle = ECCV,
    year = {2022}
}

@article{yao2021filip,
  title={FILIP: Fine-grained interactive language-image pre-training},
  author={Yao, Lewei and Huang, Runhui and Hou, Lu and Lu, Guansong and Niu, Minzhe and Xu, Hang and Liang, Xiaodan and Li, Zhenguo and Jiang, Xin and Xu, Chunjing},
  journal={arXiv preprint arXiv:2111.07783},
  year={2021}
}

@inproceedings{chen2023revisiting,
  title={Revisiting multimodal representation in contrastive learning: from patch and token embeddings to finite discrete tokens},
  author={Chen, Yuxiao and Yuan, Jianbo and Tian, Yu and Geng, Shijie and Li, Xinyu and Zhou, Ding and Metaxas, Dimitris N and Yang, Hongxia},
  booktitle=CVPR,
  pages={15095--15104},
  year={2023}
}

@inproceedings{li2022blip,
  title={BLIP: Bootstrapping language-image pre-training for unified vision-language understanding and generation},
  author={Li, Junnan and Li, Dongxu and Xiong, Caiming and Hoi, Steven},
  booktitle=ICML,
  pages={12888--12900},
  year={2022},
  organization={PMLR}
}

@article{li2022fine,
  title={Fine-grained semantically aligned vision-language pre-training},
  author={Li, Juncheng and He, Xin and Wei, Longhui and Qian, Long and Zhu, Linchao and Xie, Lingxi and Zhuang, Yueting and Tian, Qi and Tang, Siliang},
  journal={Advances in neural information processing systems},
  volume={35},
  pages={7290--7303},
  year={2022}
}

@inproceedings{goktepeecomapper,
  title={EcoMapper: Generative Modeling for Climate-Aware Satellite Imagery},
  author={Goktepe, Muhammed and hossein Shamseddin, Amir and Uysal, Erencan and Monteagudo, Javier Muinelo and Drees, Lukas and Toker, Aysim and Asseng, Senthold and von Bloh, Malte},
  booktitle={Forty-second International Conference on Machine Learning},
  year={2025}
}

@inproceedings{zhang2023adding,
  title={Adding conditional control to text-to-image diffusion models},
  author={Zhang, Lvmin and Rao, Anyi and Agrawala, Maneesh},
  booktitle={Proceedings of the IEEE/CVF International Conference on Computer Vision},
  pages={3836--3847},
  year={2023}
}

@article{kirillov2023segment,
  title={Segment Anything},
  author={Kirillov, Alexander and Mintun, Eric and Ravi, Nikhila and Mao, Hanzi and Rolland, Chloe and Gustafson, Laura and Xiao, Tete and Whitehead, Spencer and Berg, Alexander C and Lo, Wan-Yen and others},
  journal={arXiv preprint arXiv:2304.02643},
  year={2023}
}

@article{khanna2023diffusionsat,
  title={DiffusionSat: A Generative Foundation Model for Satellite Imagery},
  author={Khanna, Samar and Liu, Patrick and Zhou, Linqi and Meng, Chenlin and Rombach, Robin and Burke, Marshall and Lobell, David and Ermon, Stefano},
  journal={arXiv preprint arXiv:2312.03606},
  year={2023}
}

@article{he2021spatial,
  title={Spatial-temporal super-resolution of satellite imagery via conditional pixel synthesis},
  author={He, Yutong and Wang, Dingjie and Lai, Nicholas and Zhang, William and Meng, Chenlin and Burke, Marshall and Lobell, David and Ermon, Stefano},
  journal={Advances in Neural Information Processing Systems},
  volume={34},
  pages={27903--27915},
  year={2021}
}

@article{liu2024visual,
  title={Visual instruction tuning},
  author={Liu, Haotian and Li, Chunyuan and Wu, Qingyang and Lee, Yong Jae},
  journal={Advances in neural information processing systems},
  volume={36},
  year={2024}
}

@article{jarvelin2002cumulated,
  title={Cumulated gain-based evaluation of IR techniques},
  author={J{\"a}rvelin, Kalervo and Kek{\"a}l{\"a}inen, Jaana},
  journal={ACM Transactions on Information Systems (TOIS)},
  volume={20},
  number={4},
  pages={422--446},
  year={2002},
  publisher={ACM New York, NY, USA}
}

@article{wang2004imagessim,
  title={Image quality assessment: from error visibility to structural similarity},
  author={Wang, Zhou and Bovik, Alan C and Sheikh, Hamid R and Simoncelli, Eero P},
  journal={IEEE transactions on image processing},
  volume={13},
  number={4},
  pages={600--612},
  year={2004},
  publisher={IEEE}
}

@article{heusel2017gans,
  title={GANs trained by a two time-scale update rule converge to a local nash equilibrium},
  author={Heusel, Martin and Ramsauer, Hubert and Unterthiner, Thomas and Nessler, Bernhard and Hochreiter, Sepp},
  journal={Advances in neural information processing systems},
  volume={30},
  year={2017}
}

@article{zhang2019bertscore,
  title={Bertscore: Evaluating text generation with bert},
  author={Zhang, Tianyi and Kishore, Varsha and Wu, Felix and Weinberger, Kilian Q and Artzi, Yoav},
  journal={arXiv preprint arXiv:1904.09675},
  year={2019}
}

@inproceedings{jia2021scaling,
  title={Scaling up visual and vision-language representation learning with noisy text supervision},
  author={Jia, Chao and Yang, Yinfei and Xia, Ye and Chen, Yi-Ting and Parekh, Zarana and Pham, Hieu and Le, Quoc and Sung, Yun-Hsuan and Li, Zhen and Duerig, Tom},
  booktitle={International conference on machine learning},
  pages={4904--4916},
  year={2021},
  organization={PMLR}
}

@inproceedings{li2025segearth,
  title={SegEarth-OV: Towards training-free open-vocabulary segmentation for remote sensing images},
  author={Li, Kaiyu and Liu, Ruixun and Cao, Xiangyong and Bai, Xueru and Zhou, Feng and Meng, Deyu and Wang, Zhi},
  booktitle={Proceedings of the Computer Vision and Pattern Recognition Conference},
  pages={10545--10556},
  year={2025}
}

@inproceedings{zhong2022regionclip,
  title={RegionCLIP: Region-based language-image pretraining},
  author={Zhong, Yiwu and Yang, Jianwei and Zhang, Pengchuan and Li, Chunyuan and Codella, Noel and Li, Liunian Harold and Zhou, Luowei and Dai, Xiyang and Yuan, Lu and Li, Yin and others},
  booktitle={Proceedings of the IEEE/CVF conference on computer vision and pattern recognition},
  pages={16793--16803},
  year={2022}
}

@article{zeng2024maskclip++,
  title={MaskCLIP++: A mask-based clip fine-tuning framework for open-vocabulary image segmentation},
  author={Zeng, Quan-Sheng and Li, Yunheng and Zhou, Daquan and Li, Guanbin and Hou, Qibin and Cheng, Ming-Ming},
  year={2024}
}

@article{jing2024fineclip,
  title={FineCLIP: Self-distilled region-based clip for better fine-grained understanding},
  author={Jing, Dong and He, Xiaolong and Luo, Yutian and Fei, Nanyi and Wei, Wei and Zhao, Huiwen and Lu, Zhiwu and others},
  journal={Advances in Neural Information Processing Systems},
  volume={37},
  pages={27896--27918},
  year={2024}
}

@inproceedings{lan2024clearclip,
  title={ClearCLIP: Decomposing clip representations for dense vision-language inference},
  author={Lan, Mengcheng and Chen, Chaofeng and Ke, Yiping and Wang, Xinjiang and Feng, Litong and Zhang, Wayne},
  booktitle={European Conference on Computer Vision},
  pages={143--160},
  year={2024},
  organization={Springer}
}

@article{pang2023ssrgan,
  title     = {The Use of a Stable Super-Resolution Generative Adversarial Network (SSRGAN) on Remote Sensing Images},
  author    = {Pang, Boyu and Zhao, Siwei and Liu, Yinnian},
  journal   = {Remote Sensing},
  volume    = {15},
  number    = {20},
  pages     = {5064},
  year      = {2023},
  publisher = {MDPI},
  doi       = {10.3390/rs15205064},
  url       = {https://doi.org/10.3390/rs15205064}
}

@article{mahara2024multispectral,
  title     = {Multispectral Band-Aware Generation of Satellite Images across Domains Using Generative Adversarial Networks and Contrastive Learning},
  author    = {Mahara, Arpan and Rishe, Naphtali},
  journal   = {Remote Sensing},
  volume    = {16},
  number    = {7},
  pages     = {1154},
  year      = {2024},
  publisher = {MDPI},
  doi       = {10.3390/rs16071154},
  url       = {https://doi.org/10.3390/rs16071154}
}

@inproceedings{radford2021learning,
  title={Learning transferable visual models from natural language supervision},
  author={Radford, Alec and Kim, Jong Wook and Hallacy, Chris and Ramesh, Aditya and Goh, Gabriel and Agarwal, Sandhini and Sastry, Girish and Askell, Amanda and Mishkin, Pamela and Clark, Jack and others},
  booktitle={International conference on machine learning},
  pages={8748--8763},
  year={2021},
  organization={PMLR}
}

@inproceedings{lugmayr2022repaint,
  title={Repaint: Inpainting using denoising diffusion probabilistic models},
  author={Lugmayr, Andreas and Danelljan, Martin and Romero, Andres and Yu, Fisher and Timofte, Radu and Van Gool, Luc},
  booktitle={Proceedings of the IEEE/CVF conference on computer vision and pattern recognition},
  pages={11461--11471},
  year={2022}
}

@inproceedings{sastry2024geosynth,
  title={GeoSynth: Contextually-Aware High-Resolution Satellite Image Synthesis},
  author={Sastry, Srikumar and Khanal, Subash and Dhakal, Aayush and Jacobs, Nathan},
  booktitle={Proceedings of the IEEE/CVF Conference on Computer Vision and Pattern Recognition},
  pages={460--470},
  year={2024}
}

@article{zhang2023diffucd,
  title={DiffUCD: Unsupervised hyperspectral image change detection with semantic correlation diffusion model},
  author={Zhang, Xiangrong and Tian, Shunli and Wang, Guanchun and Zhou, Huiyu and Jiao, Licheng},
  journal={arXiv preprint arXiv:2305.12410},
  year={2023}
}

@inproceedings{benidir2025change,
  title={The Change You Want To Detect: Semantic Change Detection In Earth Observation With Hybrid Data Generationf},
  author={Benidir, Yanis and Gonthier, Nicolas and Mallet, Cl{\'e}ment},
  booktitle={Proceedings of the Computer Vision and Pattern Recognition Conference},
  pages={2204--2214},
  year={2025}
}

@article{yu2024metaearth,
  title={Metaearth: A generative foundation model for global-scale remote sensing image generation},
  author={Yu, Zhiping and Liu, Chenyang and Liu, Liqin and Shi, Zhenwei and Zou, Zhengxia},
  journal={IEEE Transactions on Pattern Analysis and Machine Intelligence},
  year={2024},
  publisher={IEEE}
}

@article{liu2024remoteclip,
  author       = {Fan Liu and
                  Delong Chen and
                  Zhangqingyun Guan and
                  Xiaocong Zhou and
                  Jiale Zhu and
                  Qiaolin Ye and
                  Liyong Fu and
                  Jun Zhou},
  title        = {RemoteCLIP: {A} Vision Language Foundation Model for Remote Sensing},
  journal      = {{IEEE} Transactions on Geoscience and Remote Sensing},
  volume       = {62},
  pages        = {1--16},
  year         = {2024},
  url          = {https://doi.org/10.1109/TGRS.2024.3390838},
  doi          = {10.1109/TGRS.2024.3390838},
}

@misc{yao2025RemoteSAM,
      title={RemoteSAM: Towards Segment Anything for Earth Observation}, 
      author={Liang Yao and Fan Liu and Delong Chen and Chuanyi Zhang and Yijun Wang and Ziyun Chen and Wei Xu and Shimin Di and Yuhui Zheng},
journal      = {Proceedings of the 33rd ACM International Conference on Multimedia},
      year={2025},
      eprint={2505.18022},
      archivePrefix={arXiv},
      primaryClass={cs.CV},
      url={https://arxiv.org/abs/2505.18022}, 
}

@article{wang2023cloud,
  title={Cloud removal with SAR-optical data fusion using a unified spatial--spectral residual network},
  author={Wang, Yuxi and Zhang, Bing and Zhang, Wenjuan and Hong, Danfeng and Zhao, Bin and Li, Zhen},
  journal={IEEE Transactions on Geoscience and Remote Sensing},
  volume={62},
  pages={1--20},
  year={2023},
  publisher={IEEE}
}

@InProceedings{Han_2025_CVPR,
    author    = {Han, Tengda and Gokay, Dilara and Heyward, Joseph and Zhang, Chuhan and Zoran, Daniel and Patraucean, Viorica and Carreira, Joao and Damen, Dima and Zisserman, Andrew},
    title     = {Learning from Streaming Video with Orthogonal Gradients},
    booktitle = {Proceedings of the Computer Vision and Pattern Recognition Conference (CVPR)},
    month     = {June},
    year      = {2025},
    pages     = {13651-13660}
}

@misc{google_microsoft_open_buildings,
  author       = {{VIDA}},
  title        = {Google-Microsoft Open Buildings - combined by VIDA},
  howpublished = {\url{https://beta.source.coop/repositories/vida/google-microsoft-open-buildings}},
  note         = {Accessed: 2025-05-18}
}

@article{baisar2optical,
  author={Bai, Xinyu and Pu, Xinyang and Xu, Feng},
  journal={IEEE Geoscience and Remote Sensing Letters}, 
  title={Conditional Diffusion for SAR to Optical Image Translation}, 
  year={2023},
  volume={},
  number={},
  pages={1-1},
  doi={10.1109/LGRS.2023.3337143}
}
}

\clearpage
\appendix
\maketitlesupplementary

\section{Applications}

\begin{figure*}[t]
  \centering
  \includegraphics[width=\textwidth]{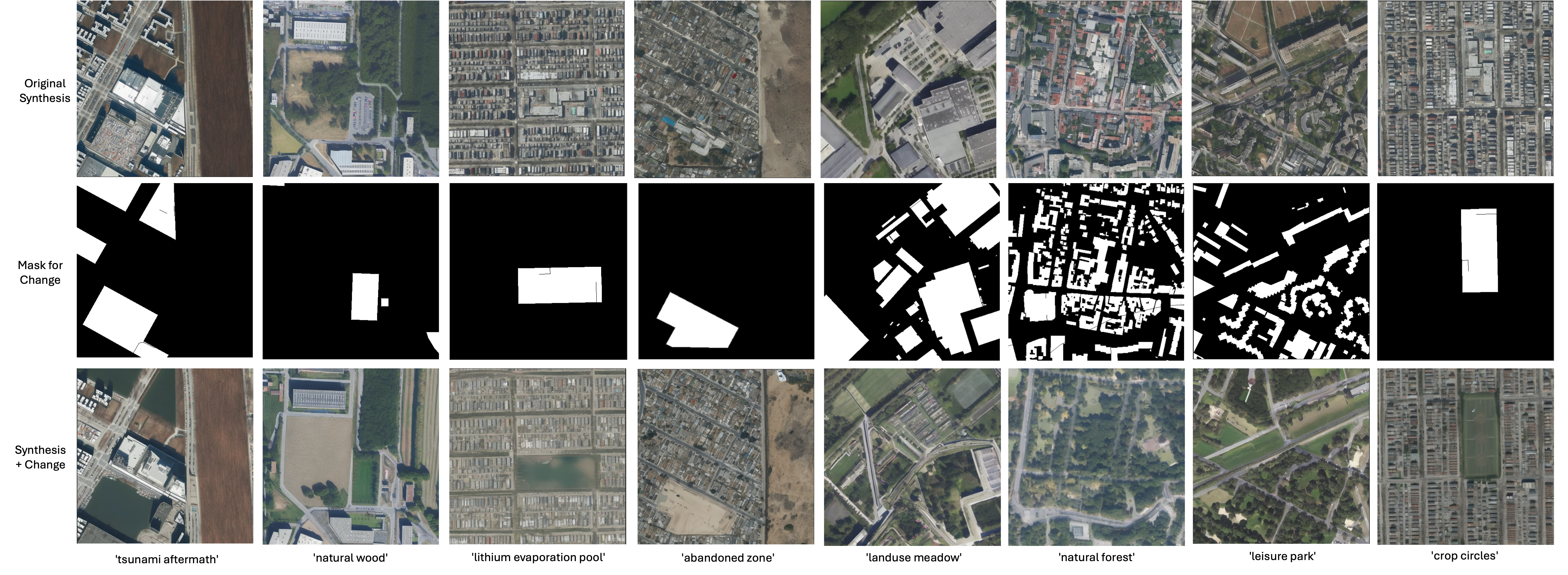}
  \caption{
    Qualitative evaluations on out-of-distribution text incorporated conditions, and purposeful edits. Each set shows: (top) original synthesized image, (middle) the injected control mask corresponding to the semantic change, and (bottom) the synthesized output conditioned on (below) the natural language description used for conditioning.
  }
  \label{fig:funky_qual_evals}
\end{figure*}

We evaluate the utility of our generated imagery by running SegEarth-OV~\cite{li2025segearth}, a state-of-the-art open-vocabulary remote sensing segmentation model, on synthetic images created from structured OSM tags using our VectorSynth-COSA model. We also run the segmentation model on the grounded satellite imagery, and GeoSynth-OSM~\cite{sastry2024geosynth} for comparison.  We define a subset of categories in our data that represent different land uses, buildings, and road types. We use segmentation accuracy to measure how well the generated image matches the given polygon labels, with higher accuracy meaning the generated image shows strong pixel-level fidelity to the class. 

We consistently outperform GeoSynth-OSM across all categories, with particularly strong gains in fine-grained classes such as road types and distinct building uses. As shown in Table~\ref{tab:segearth_hierarchy}, our method achieves competitive results compared to real satellite imagery, and in several categories, such as land use residential, and natural regions, performance even surpasses that of real images. This indicates that our model generates semantically faithful scenes that align well with downstream open-vocabulary segmentation tasks. While challenging categories like industrial areas remain difficult due to visual ambiguity, our results demonstrate that our pretraining alignment and generation pipeline yields more spatially and semantically precise synthetic outputs.

We further perform some qualitative evaluations in Figure~\ref{fig:funky_qual_evals} on out of OSM distribution text prompts. We see that generated outputs adhere to spatial and semantic constraints.

\begin{table}[t]
\renewcommand{\arraystretch}{0.95}
\setlength{\tabcolsep}{6pt}
\centering
\small
\begin{tabular}{lccc}
\toprule
\textbf{Class} & \textbf{GeoSynth} & \textbf{VectorSynth} & \textbf{Original} \\
\midrule
place              & 79.54 & 81.26 & 81.81 \\
\quad natural region   & 25.26 & 26.04 & 25.55 \\
building           & 19.52 & 32.03 & 32.43 \\
\quad industrial       & 4.15  & 18.23 & 36.19 \\
\quad apartments       & 5.00  & 14.95 & 22.92 \\
\quad school           & 0.50  & 11.39 & 20.69 \\
landuse            & 44.62 & 55.90 & 55.06 \\
\quad residential      & 44.62 & 55.90 & 55.06 \\
\quad farmland         & 2.35  & 16.80 & 55.17 \\
\quad forest           & 12.39 & 28.12 & 36.47 \\
sport              & 4.86  & 15.51 & 26.65 \\
railway            & 2.93  & 13.51 & 42.04 \\
\bottomrule
\end{tabular}
\caption{Segmentation accuracy (\%) for parent and child classes in OSM tags using SegEarth-OV. Child classes are indented under their parent. We compare generated images from GeoSynth and VectorSynth, along with the original satellite image.}
\label{tab:segearth_hierarchy}
\end{table}

\section{Data}
\begin{figure*}
\centering
\includegraphics[width=0.9\linewidth]{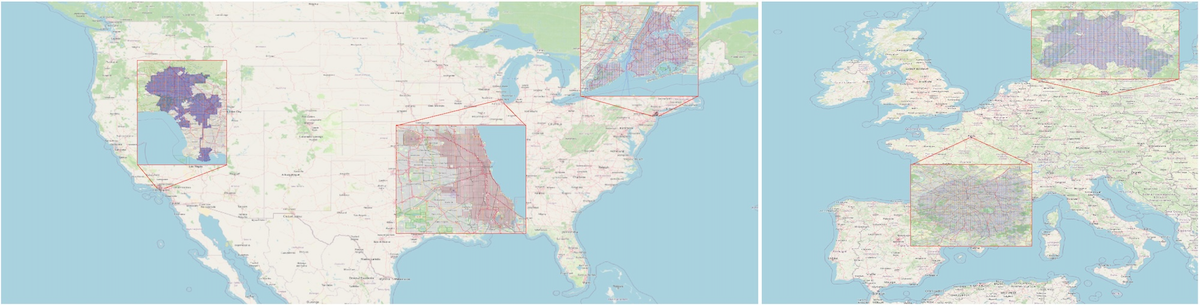}
\caption{Geographic coverage of the dataset across five major cities: Los Angeles, New York City, Paris, Berlin, and Chicago. Each city includes training, validation, and held-out tiles, except for Chicago, which is fully held out and used only for testing. Training and validation tiles are shown in blue; Chicago test tiles are shown in red.}
\label{fig:data_coverage}
\end{figure*}

As seen in Figure~\ref{fig:data_coverage}, we densely sampled Los Angeles, New York City, Paris, and Berlin. These cities were chosen as they represent different urban planning styles: Los Angeles exemplifies low-density horizontal sprawl, New York is defined by verticality and a rigid grid system, Paris features radial layouts and dense historical cores, and Berlin reflects a blend of post-war reconstruction and structured zoning. Chicago is used as an out-of-space test city. In addition, we conduct one additional experiment on generating OSM annotations using Sydney, Australia.

\begin{figure}[t]
  \centering
  \includegraphics[width=0.9\linewidth]{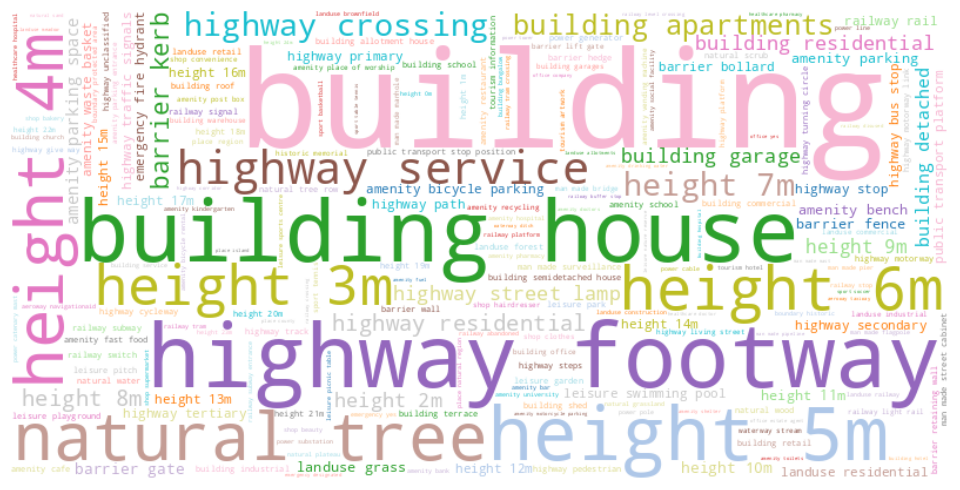}
  \caption{
    Word cloud of the most frequent OSM tags in the dataset. Font size reflects frequency across all tile-level tag lists.
    Common tags include urban structure types (e.g., \texttt{building residential}, \texttt{highway primary}), land use (\texttt{land use commercial}, \texttt{park}), and 3D features such as \texttt{height}.
  }
  \label{fig:tag_wordcloud}
\end{figure}

\begin{figure}[t]
  \centering
  \includegraphics[width=0.9\linewidth]{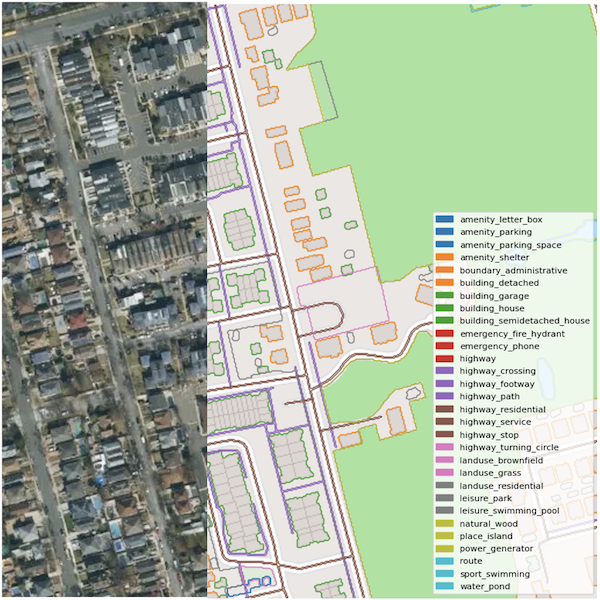}
  \caption{Overlay of OSM-derived tags on top of satellite imagery. Each region is annotated with semantically meaningful labels (e.g., \texttt{building residential}, \texttt{land use park}), showcasing the compositional richness and spatial precision of the dataset.}
  \label{fig:tag_overlay}
\end{figure}

We visualize the tag distribution in our training data as seen in Figure~\ref{fig:tag_wordcloud}. Through the overlay visualization in Figure~\ref{fig:tag_overlay}, we see that there is strong spatial grounding in the dataset. 

\section{COSA}

\begin{figure*}
\centering
\includegraphics[width=1\linewidth]{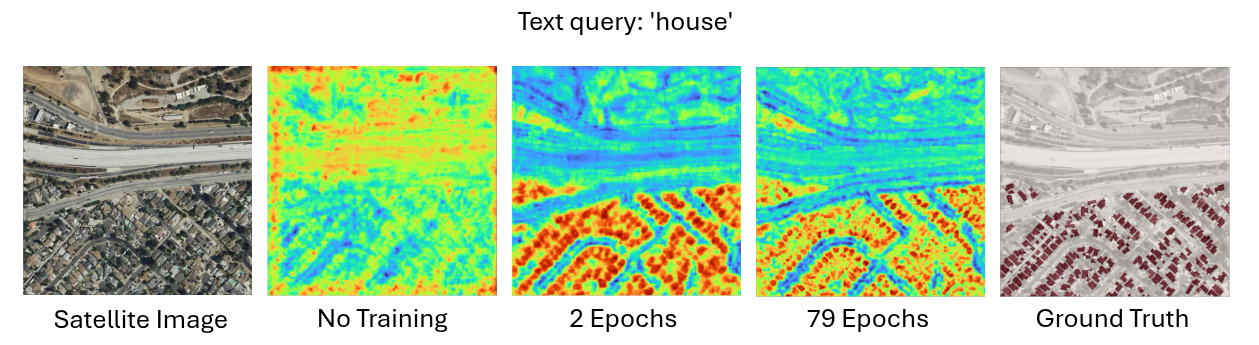}
\caption{Similarity heatmaps for a satellite image given the text query `house' inferred from COSA with no training, 2 epochs of training, and 79 epochs of training.}
\label{fig:tag_sim_heatmap}
\end{figure*}

\begin{figure*}
\centering
\includegraphics[width=0.95\linewidth]{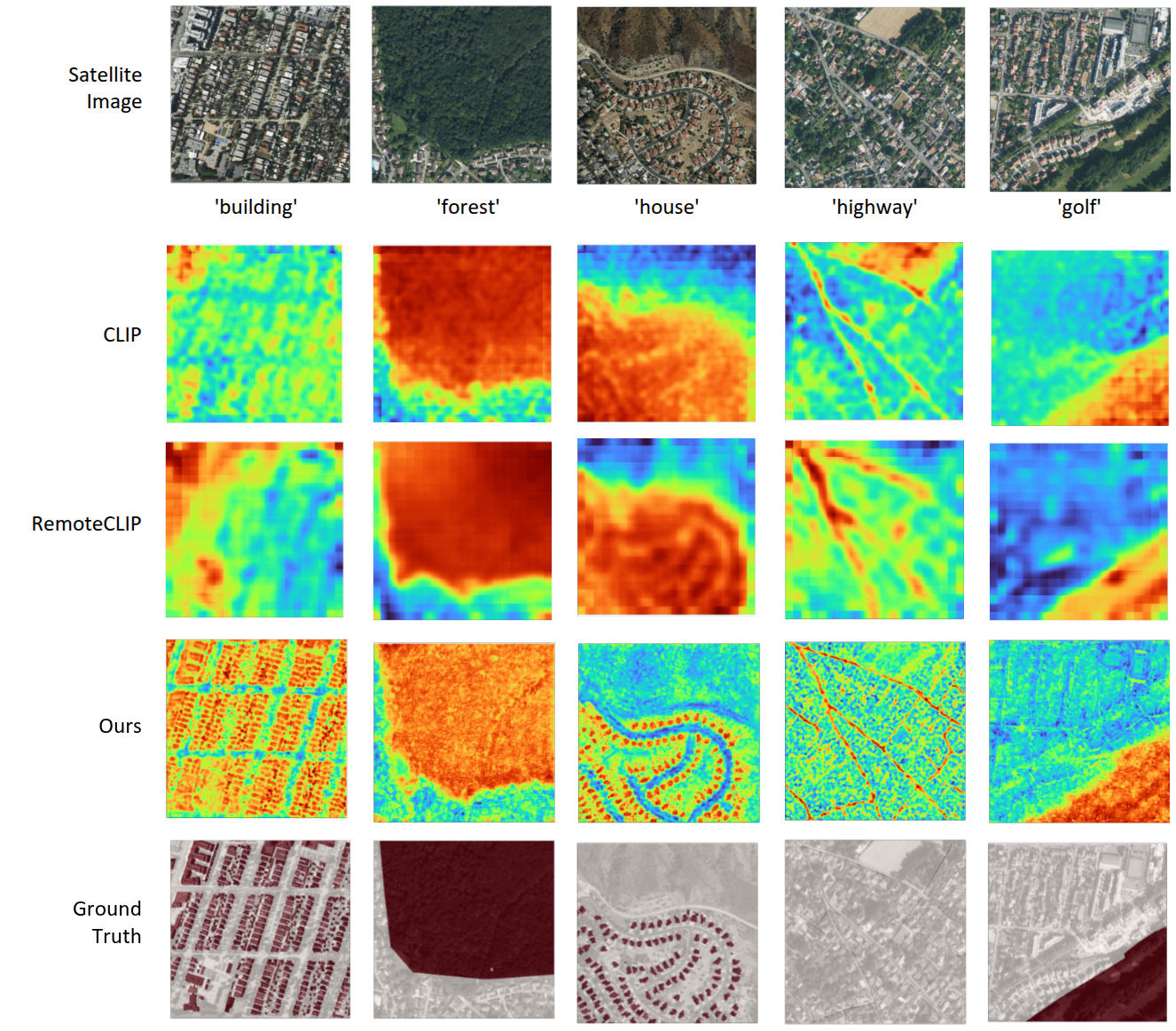}
\caption{Similarity heatmaps given text queries comparing CLIP, RemoteCLIP, and our approach—COSA. Taking inspiration from~\cite{liu2024remoteclip}, we use a sliding window inference approach to show high-resolution similarity heatmaps for CLIP and RemoteCLIP.}
\label{fig:tag_sim_heatmap}
\end{figure*}

\subsection{Architecture Details}

\paragraph{Image Encoder.}

Our image encoder is built on top of SatlasNet~\cite{bastani2023satlaspretrain}. SatlasNet is pretrained on high-resolution aerial imagery, consistent with our dataset, using a Swin-V2 backbone followed by a feature pyramid network (FPN) resulting in multi-scale feature maps of varying resolution. Following the FPN, our image encoder interpolates and concatenates the multi-scale feature maps, then passes the result through a learnable MLP adapter network to align the embedding dimension of the text encoder. Our adapter network consists of sequential 1×1 convolutional layers with ReLU activations and Batch Normalization, following a Conv2d–ReLU–BN–Conv2d–ReLU–BN structure. To encourage high resolution vision-language alignment, we freeze the Swin-V2 backbone and let the feature pyramid network, and adapter network be learnable.




\subsection{Training and Implementation Details}
\paragraph{Polygon Sampling.}

As the number of polygon-text pairs varies within a minibatch, contrastive sampling size can also fluctuate. To address this variability, we use a combination of minibatch size $B$ and number of sampled polygon-text pairs $K$ such that, with at least 95\% confidence, the sampled $K$ pairs is reached. In cases where $K$ is not reached, we sample all polygon-text pairs within the batch. If the same multi-tag composition is in multiple images, the polygon pair is randomly chosen from one of the corresponding images. Our setup naturally introduces both \textit{intra-image} and \textit{inter-image} negatives, encouraging distinction between semantically similar polygons-text pairs within the same image and across different images. In addition, for each contrastive pair, we sub-sample tag words in the multi-tag compositions during training to provide better generalization to varying text queries.

\paragraph{Optimization With Orthogonal Gradients.}
Due to the spatial nature of satellite images, polygon-text pairs often exhibit strong feature correlation, both within the same image and across images in the batch. This is especially true in urban areas, where OSM tag distributions and architectural layouts likely follow recurring spatial patterns (e.g., residential blocks, road grids, building clusters). Such correlations may limit learning efficiency and lead to gradient directions with poor diversity. For this reason, we implement orthogonal gradients~\cite{Han_2025_CVPR}, an optimization technique designed to promote diversity by projecting updates onto the gradients orthogonal component. This approach has shown effectiveness in data domains such as sequential video frames, where the data is highly correlated. Specifically, we implement the \textit{Orthogonal AdamW} variant implemented based on ~\cite{Han_2025_CVPR}. Orthogonal AdamW introduces an additional term controlled by a hyperparameter $\beta_{\text{ort}}$, set to $0.9$ in our experiments.



\paragraph{Training Details.}
We train our model using the AdamW optimizer with a learning rate of $1\text{e}{-4}$, $\beta$ values of $(0.9, 0.98)$, $\epsilon=1\text{e}{-6}$, and a weight decay of $0.01$. We use a cosine annealing warm restart learning rate scheduler with an initial cycle length of $T_0 = 20$ epochs. We train until early stopping with a minibatch size of $B = 6$ satellite images per GPU. Each batch includes $K = 128$ sampled polygon-tag pairs, drawn across the minibatch. Training typically ended around 80 epochs. We initialize the logit scale temperature parameter as $\log(1/0.07)$ and learn it during training. To avoid numerical instability, we clamp the logit scale to a maximum of $\log(100)$. To ensure reproducibility, we set all random seeds to 42 and disabled CuDNN benchmarking. All experiments are run on two GeForce RTX 4090 GPUs (24GB). 

\section{VectorSynth Controls}

\begin{figure*}
\centering
\includegraphics[width=0.8\linewidth]{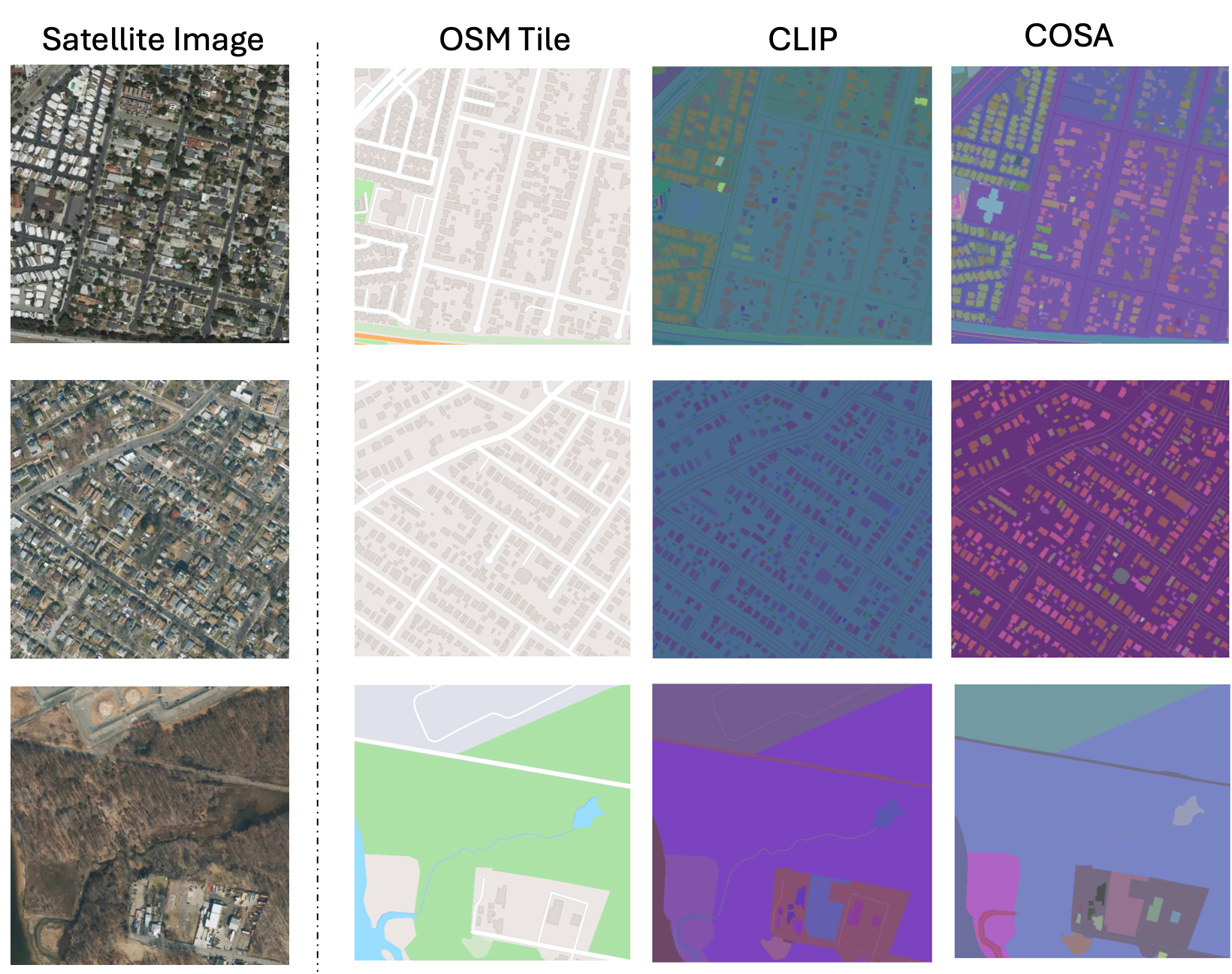}
\caption{
Qualitative comparison of control signals. OSM tiles provide coarse structural priors but lack semantic detail. Text-based maps offer richer semantics, while COSA maps show sharper object boundaries and better spatial grounding—especially in distinguishing urban features like residential and commercial buildings. These improvements stem from aligning OSM tags with satellite imagery during pretraining.
}
\label{fig:control_image_comparison}
\end{figure*}

A qualitative comparison of the different control signals is provided in Figure~\ref{fig:control_image_comparison}. Visually, we observe that OSM tiles provide a high-level structural prior but lack semantic richness. While text-based pixel-level control maps introduce more diverse semantic information, our COSA control maps exhibit sharper transitions between objects, reflecting stronger inter-tag contrast and improved spatial grounding. This is especially evident in the fine-grained delineation of urban features. For example, residential and commercial buildings, as well as differences in heights of buildings, appear more homogeneous in CLIP maps, but are more distinctly separated in COSA. These improvements result from aligning OSM tag semantics with satellite imagery during pretraining, leading to control signals that are both semantically expressive and spatially localized.

\section{Dealing with Sparsity}

Geographic annotation datasets often suffer from inherent sparsity, where comprehensive polygon coverage is unavailable across all spatial regions. This sparsity presents significant challenges during both training and inference, as models must generate plausible geographic content even when provided with incomplete or limited control signals. We address this fundamental limitation through two complementary approaches: progressive masking during training and automated annotation enhancement using vision-language models.

\subsection{Progressive Masking for Sparse Control Adaptation}

\begin{figure}[t]
  \centering
  \includegraphics[width=0.9\linewidth]{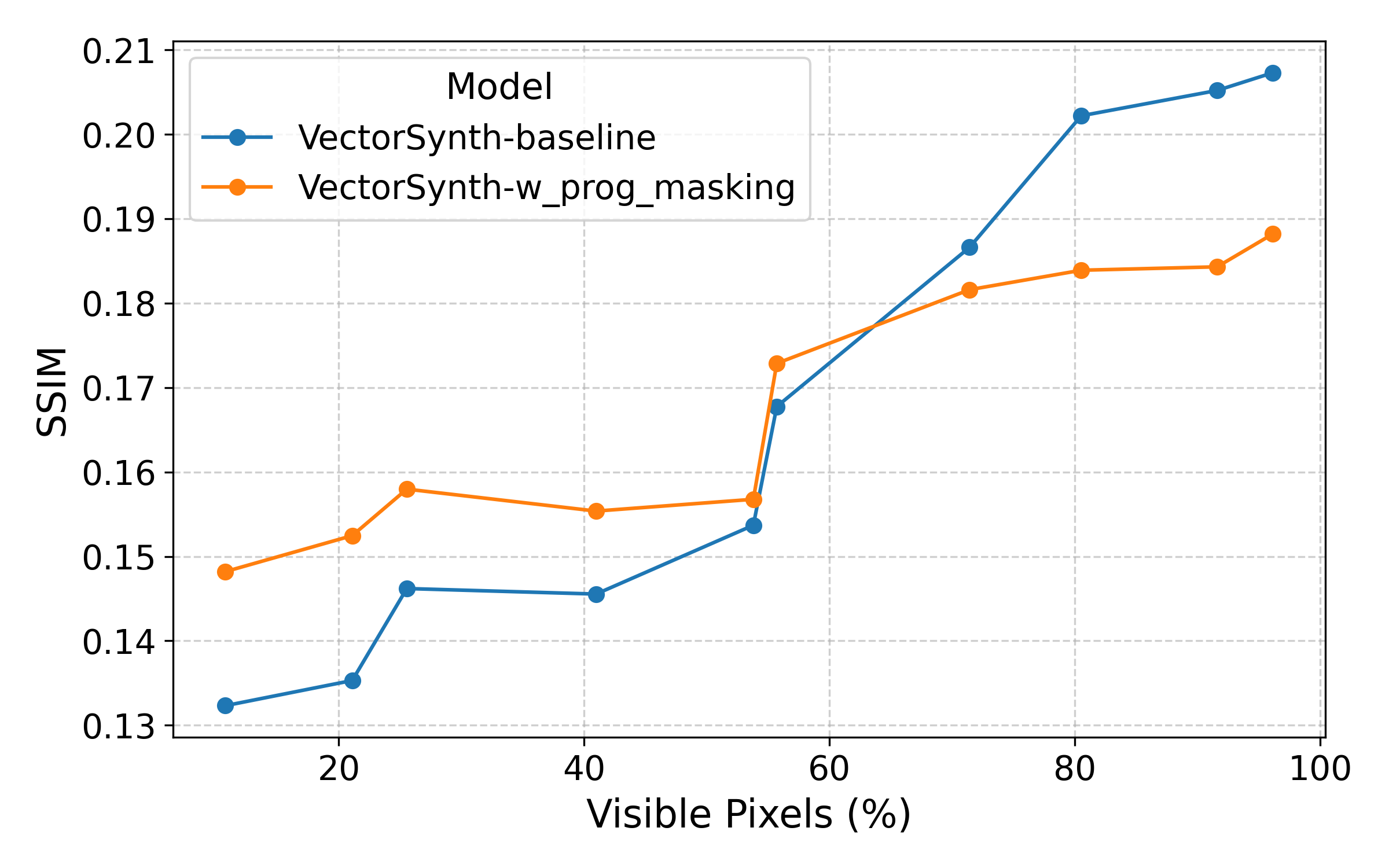}
  \caption{
    SSIM performance versus polygon coverage. Progressive masking (orange) outperforms baseline (blue) below 60\% coverage but underperforms at dense coverage above 80\%.
  }
  \label{fig:progressive_masking}
\end{figure}

To enable robust performance under sparse annotation conditions, we use a progressive masking training strategy that gradually reduces polygon coverage throughout the training process. This approach trains the model to effectively hallucinate plausible geographic features when given increasingly sparse control inputs.

Our progressive masking scheme linearly increases the proportion of masked polygons over training iterations (100\% to 30\%). This curriculum learning approach allows the model to first establish strong associations between dense annotations and corresponding geographic features, then gradually adapt to scenarios with limited supervisory information.

The progressive masking strategy demonstrates clear benefits for sparse control scenarios. As illustrated in Figure~\ref{fig:progressive_masking}, models trained with this approach exhibit improved robustness when polygon coverage falls below 60\% of the image area. However, we observe a trade-off in performance: while the progressively masked model excels with sparse controls, it slightly underperforms compared to the baseline model when provided with very dense annotation coverage. This behavior aligns with our training objective, as the model learns to rely less heavily on comprehensive annotation signals.

\subsection{Automated Annotation Enhancement via COSA VLM}

\begin{table*}[t]
\centering
\small
\begin{tabular}{lccc}
\toprule
\textbf{Model} & \textbf{FID} $\downarrow$ & \textbf{SSIM} $\uparrow$ & \textbf{PSNR} $\uparrow$ \\
\midrule
GeoSynth & 170.25 & 0.18 & 12.16 \\
VectorSynth & 177.17 & 0.17 & 11.99 \\
VectorSynth (w/ generated tags) & 154.13 & 0.18 & 12.11 \\
\bottomrule
\end{tabular}
\caption{Comparison of FID, SSIM, and PSNR of satellite imagery across Sydney, Australia}
\label{tab:sydney_eval}
\end{table*}

To further address annotation sparsity, we leverage our COSA vision-language model (VLM) to automatically generate additional semantic annotations for sparse regions. This approach combines the Segment Anything Model (SAM)~\cite{kirillov2023segment} for mask generation with our specialized COSA VLM for polygon-text retrieval, creating a pipeline that densifies sparse annotations with contextually appropriate semantic labels.

The annotation enhancement pipeline operates in three stages. First, we apply SAM~\cite{kirillov2023segment} to the input satellite imagery to generate comprehensive segmentation masks covering all visible geographic features. Next, we utilize our COSA VLM to perform polygon-retrieval, generating semantically grounded text descriptions for each SAM-generated mask. These automatically generated text annotations are then integrated with existing sparse annotations to provide richer control signals during generation.

We evaluate this approach on an out-of-distribution dataset featuring high-resolution imagery of Sydney, Australia. Sydney's harbor-centric development and organic street patterns contrast with our training data from NYC, LA, Berlin, and Paris, which feature more geometric grids and radial planning structures. We compare three generation approaches: VectorSynth using only available OpenStreetMap (OSM)~\cite{haklay2008osm} tags without filtering, the baseline GeoSynth model, and VectorSynth enhanced with SAM + COSA VLM annotations. We note that we do not filter the coverage of the dataset; therefore, the OSM tags are very sparse, and many images do not contain any OSM tag information.

In Table~\ref{tab:sydney_eval}, we see that using our text generation pipeline improves upon strictly using the OSM tags, and outperforms other baselines. Our experimental results demonstrate that the automated annotation enhancement pipeline can be an effective way to mitigates sparsity limitations and generate data that is useful for our vectorsynth generation.

The combination of progressive masking training and automated annotation enhancement provides a comprehensive solution to the sparsity challenge in geographic image synthesis. While progressive masking enables the model to perform well with inherently sparse controls, the COSA VLM pipeline allows us to artificially densify annotations when computational resources permit, achieving the best of both sparse and dense control paradigms.




\end{document}